\pdfoutput=1

\documentclass[11pt]{article}

\usepackage[final]{coling}
\usepackage{times}
\usepackage{latexsym}
\usepackage{float}
\usepackage{placeins}

\usepackage{colortbl}
\usepackage{xcolor} 
\usepackage{hyperref}
\usepackage{url}
\usepackage{booktabs}       
\usepackage{amsfonts}       
\usepackage{nicefrac}       
\usepackage{microtype}      
\usepackage{wrapfig}
\usepackage{latexsym}
\usepackage{amsfonts}
\usepackage{amsmath}
\usepackage{siunitx}
\usepackage{dcolumn}
\usepackage{arydshln}
\usepackage{multirow}
\usepackage{booktabs}
\usepackage{graphicx}
\usepackage{algorithm, algpseudocode}
\usepackage{graphicx}
\usepackage{verbatim}
\usepackage{stmaryrd}
\usepackage{trimclip}
\usepackage{subcaption}
\usepackage{url}
\usepackage{xspace}
\usepackage[normalem]{ulem}
\usepackage[most]{tcolorbox}
\usepackage{tabularx}
\usepackage{fontawesome5}
\usepackage{amssymb}
\usepackage{pifont}
\usepackage{times}
\usepackage{latexsym}
\usepackage{amsmath, array, multirow, graphicx}
\usepackage{booktabs}
\usepackage{multirow}
\usepackage{array}
\usepackage[T1]{fontenc}
\usepackage{amsmath}

\usepackage[utf8]{inputenc}

\usepackage{microtype}

\usepackage{inconsolata}
\usepackage{amssymb}

\def\modelname{LLaVAR-2}

\title{A High-Quality Text-Rich Image Instruction Tuning Dataset \\ via Hybrid Instruction Generation}

\author{
    Shijie Zhou\textsuperscript{1},
    Ruiyi Zhang\textsuperscript{2}\footnotemark[2], 
    Yufan Zhou\textsuperscript{2},
    Changyou Chen\textsuperscript{1}
    \\ 
    \textsuperscript{1}University at Buffalo~~~~~~~~~
    \textsuperscript{2}Adobe Research
}
\begin{document}
\maketitle
\renewcommand{\thefootnote}{\fnsymbol{footnote}}
\footnotetext[2]{Corresponding Author}
\renewcommand*{\thefootnote}{\arabic{footnote}}
\begin{abstract}
Large multimodal models still struggle with text-rich images because of inadequate training data. Self-Instruct provides an annotation-free way for generating instruction data, but its quality is poor, as multimodal alignment remains a hurdle even for the largest models. In this work, we propose \textbf{LLaVAR-2}\footnote{Project page: \href{https://github.com/llavar/LLaVAR-2}{https://github.com/llavar/LLaVAR-2}}, to enhance multimodal alignment for text-rich images through hybrid instruction generation between human annotators and large language models. Specifically, it involves detailed image captions from human annotators, followed by the use of these annotations in tailored text prompts for GPT-4o to curate a dataset. It also implements several mechanisms to filter out low-quality data, and the resulting dataset comprises 424k \textbf{high-quality} pairs of instructions. Empirical results show that models fine-tuned on this dataset exhibit impressive enhancements over those trained with self-instruct data.
\end{abstract}
\section{Introduction}
Instruction tuning is widely used to improve the generalization and controllability of large language models (LLMs) \cite{wang2022self,ouyang2022training,zhang2023instruction} by converting unseen tasks into instruction-output pairs. Adopting such a manner, visual instruction fine-tuning~\cite{liu2023improvedllava} enables the visual reasoning capacity of multimodal large language models (MLLMs) by injecting aligned visual tokens leveraging visual encoders such as CLIP-ViT \cite{alexey2020image, radford2021learning} and DINO \cite{caron2021emerging, tong2024eyes}. However, obstacles persist in effectively handling text-centric visual tasks for MLLMs. These challenges likely arise from the underrepresentation of text-rich images in the training dataset, such as COCO \cite{lin2014microsoft} and Conceptual Captions \cite{changpinyo2021conceptual}. However, the ability to comprehend texts within images is essential for real-world applications. 
Classical datasets on text-rich images are dedicated to information extraction, such as TextVQA~\cite{singh2019towards}, TextOCR~\cite{Singh_2021_CVPR}, DocVQA~\cite{mathew2021docvqa}, and OCR-VQA~\cite{mishraICDAR19}. Recent instruction tuning datasets for text-rich images focus more on classical document images, such as Figure~\cite{kahou2017figureqa}, Chart~\cite{masry2022chartqa} , and infographics~\cite{mathew2022infographicvqa}.

To address this issue, LLaVAR \cite{zhang2023llavar} introduces noisy and GPT-4-based instruction-following data of text-rich images, utilizing OCR and caption tools to enhance textual comprehension ability. TRINS \cite{zhang2024trins} creates a text-rich image instruction dataset in a semi-automatic manner with manual annotation effects to guarantee high-quality captions. In this work, we present LLaVAR-2, a text-rich image instruction-following dataset via hybrid instruction generation between human annotations in TRINS and LLMs to improve the effectiveness of visual instruction tuning. Specifically, we enrich the collected manual captions and the QA dataset with fine-grained details and supplementary self-explain instruction data.

LLaVAR-2 consists of two parts: LLaVAR-2-Cap for global descriptive captioning on images and LLaVAR-2-VQA for visual question answering. For LLaVAR-2-Cap data collection, rather than directly applying the human-annotated captions from TRINS as answers for crafted instructions, we rewrite the caption by incorporating necessary text/visual details to get detail-enriched captions. Based on it, we construct LLaVAR-2-Cap for precise summarizing to facilitate global visual text understanding. For instruction tuning data, we introduce a novel approach that combines extractive question answering with supplementary rounds of self-explain conversations. These pairs of self-explanations serve to illuminate the rationale behind extractive answers, supported by detailed examinations of relevant image contents. Data with extra pairs of self-explaining bolsters the localized comprehension of text-rich images, offering deeper insights and clearer connections within the visual data.
Leveraging the initial high-quality human-annotated captions, VQA and captioning data of LLaVAR-2 are generated via GPT-4o. 
Compared with TRINS \cite{zhang2024trins}, LLaVAR-2 shows better diversity in task categories, the length of instructions, and answers. To filter out low-quality data samples, we propose an automatic filtering mechanism for the multimodal instruction tuning dataset, named multimodal Instruction-following Difficulty (mIFD) score and Fact-Following Difficulty (FFD) score, to filter out incompatible extraction and self-explain pairs in VQA data.
Our contributions are as follows.
\begin{itemize}
\item We present \textbf{LLaVAR-2}, a novel dataset consisting of 42k detail-enriched captions and 382k visual question-answering data pairs, all generated automatically using GPT-4o based on human-annotated text-rich image captions. 
\item We design mIFD and FFD scores for filtering on LLaVAR-2-VQA that systematically removes irrelevant or redundant data, ensuring the dataset's high quality.
\item Beyond demonstrating the dataset's diversity through statistical visualization, we show the superiority of LLaVAR-2 by fine-tuning various base models, showing great improvement on various benchmarks.
\end{itemize}
\begin{figure*}[htp]
    \centering
    \includegraphics[width=\textwidth]{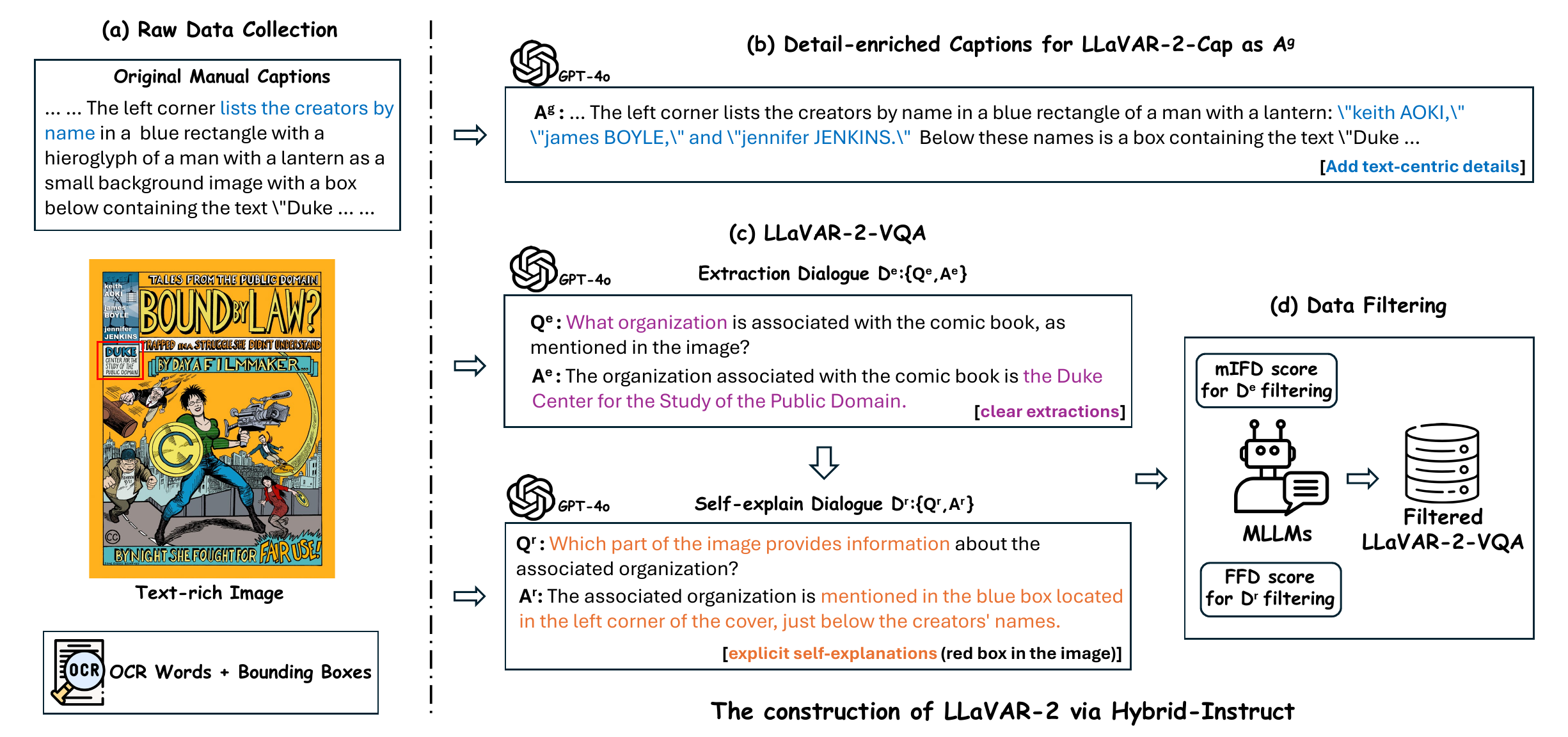}
    \caption{Overview of the data collection pipeline of Hybrid-Instruct for constructing LLaVAR-2. (a) Prompting GPT-4o with the text-rich image, manual caption, and the OCR results; (b) The Original manual caption is rewritten automatically to the detail-enriched caption, which is further used as $A^g$ to form LLaVAR-2-Cap; (c) Generated LLaVAR-2-VQA is composed of Extractive dialogue $D^e$ and Self-explain Dialogue $D^r$ which supplement explicit extraction process for $D^e$; (d) Besides, mIFD and FFD scores are proposed to filter $D^e$ and $D^r$ respectively.}
    \vspace{-0.5em}
    \label{fig:pipeline}
\end{figure*}
\section{Related Work}
\paragraph{Multimodal Large Language Models}
Recent significant success in multimodal large language models can be traced back to Flamingo~\citep{alayrac2022flamingo}, InstructBLIP~\citep{dai2023instructblip} , and LLaVA~\citep{liu2023visual}. While Flamingo and InstructBLIP form the bridge between language and vision via cross-modal attention, LLaVA-style methods transform visual representations from a standalone visual encoder into visual tokens that language models can understand. Some latest examples include Eagle~\citep{shi2024eagle}, Cambrian-1~\citep{tong2024cambrian}, LLaVA-HR~\citep{luo2024feast}, InternVL2~\citep{chen2023internvl,chen2024far}, Idefics2/3~\citep{laurenccon2024matters, laurenccon2024building}, LLaVA-OneVision~\citep{li2024llava} and Qwen2-VL~\citep{wang2024qwen2}.
Scaling up the pre-training and SFT data is a crucial part of improving MLLMs. MiniGPT-4 \citep{zhu2023minigpt} uses ChatGPT to produce data compliant with high-quality instructions, while LLaVA \citep{liu2023visual} prompts GPT-4 with image captions and boxes to similar ends. Other initiatives ~\cite{chen2023sharegpt4v, chen2024allava} prompt GPT-4V to generate more than 1 million pieces of quality data for the training of MLLMs. LLaMA-Adapter \citep{zhang2023llama, gao2023llama} synchronizes text and image features using COCO dataset inputs. InstructBLIP \citep{dai2023instructblip} has restructured 13 vision-language tasks to fit an instruction-based approach. mPLUG-Owl~\cite{ye2023ureader,ye2023mplugowl} implements multi-task instruction fine-tuning with pre-existing document datasets. VILA~\citep{lin2024vila} equips extra interleaved visual language corpus into MLLMs' pretraining. Cambrian-1~\citep{tong2024cambrian} and Idefics2~\citep{laurenccon2024matters} create large pools of instruction-tuning data containing a diverse range of visual-language tasks, such as counting, captioning, document understanding, etc.
Further research~\cite{liu2023improvedllava, liu2024llavanext, bai2023qwen, dong2024internlm, xu2024llava, luo2024feast} explores enhancing encoder resolution, leading to significant advancements in a variety of downstream applications. For a detailed examination, a comprehensive survey is provided \cite{li2023multimodal}. Despite these advances, visual-text comprehension remains a challenge for many models~\cite{liu2023hidden}.
\paragraph{Instruction-Following Data Generation} To address the lack of sufficient instruction-following data, data synthesis is commonly employed for the training of LLMs/MLLMs. Recent studies have focused on producing high-quality synthetic instruction-following data by leveraging strong and robust LLMs \cite{wang2022self,xu2023wizardlm,chen2023sharegpt4v,zhang2023llavar,zhao2023genixer,zhang2024trins}.
For instance, Self-Instruct \cite{wang2022self} boosts LLMs' instruction-following capabilities by allowing them to generate their instructional data. WizardLM \cite{xu2023wizardlm} developed Evol-Instruct to create instruction data with varying levels of complexity. In ShareGPT4V \cite{chen2023sharegpt4v}, the 1.2M image captioning data is expanded by a superb caption model trained on the initial subset. LLaVAR \cite{zhang2023llavar} targeted instruction-following data generation for text-rich images by additionally prompting LLMs with high-quality demonstrations. TRINS \cite{zhang2024trins} further improved upon LLaVAR by leveraging manually crafted high-quality image captions instead of BLIP-2 generated ones to generate instructions. 
Our proposed dataset LLaVAR-2 enriches necessary text and visual details into instruction-following data and maintains the data quality via proposed filtering scores, thus improving visual instruction tuning.
\begin{figure*}[htp]
    \centering
    \includegraphics[width=\textwidth]{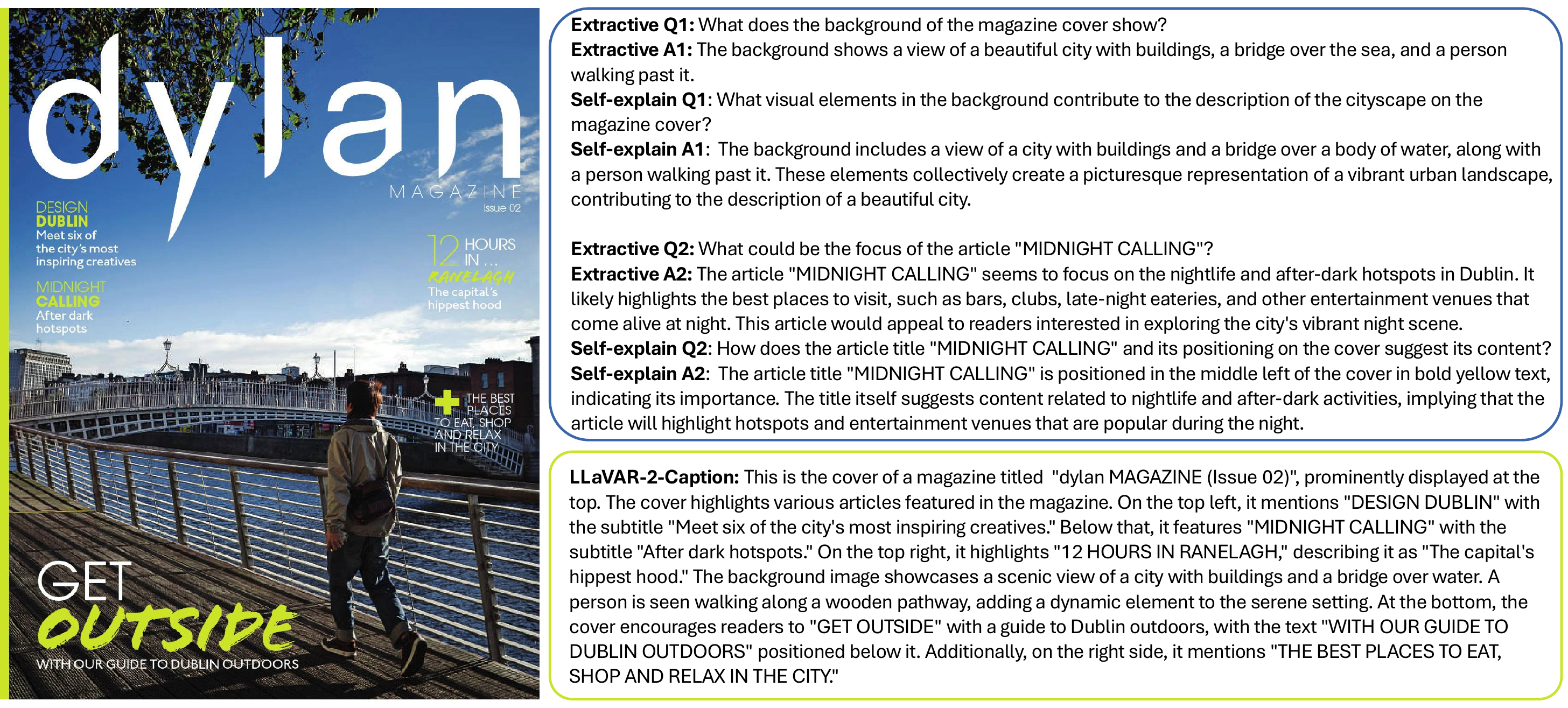}
    \vspace{-1.5em}
    \caption{An example of LLaVAR-2 dataset: a text-rich image of a magazine cover, its LLaVAR-2-Caption (Section~\ref{sec:cap}) and LLaVAR-2-VQA containing extractive and self-explain components (Section~\ref{sec:vqa}).}
    \vspace{-1em}
    \label{fig:example1}
\end{figure*}
\section{Hybrid Instruction Generation}
In Figure~\ref{fig:pipeline}, we show the data collection pipeline of \textbf{Hybrid} \textbf{Instruct}ion including the detail-enriched captions LLaVAR-2-Cap in Section~\ref{sec:cap} and the visual question answering data LLaVAR-2-VQA collection in Section~\ref{sec:vqa}. We conduct data filtering for LLaVAR-2-VQA, with the filtering process described in Section~\ref{sec:explain_filter}. Examples of LLaVAR-2-Cap and LLaVAR-2-VQA are given in Appendix~\ref{appendix:gd_example} and ~\ref{appendix:vqa_example} respectively.
\subsection{Detail-enriched Captions} \label{sec:cap} 
Manual captions in TRINS provide concrete descriptions of visual components but often offer only general and succinct summaries for text-heavy areas, lacking explicit text labels for these summaries. Thus, clear links between descriptions and corresponding text or image regions are hard to construct for MLLMs. To address this, based on OCR results, we leverage advanced GPT-4o to enrich these captions with missing details and text labels.

Specifically, to enhance global awareness of precise geometric relationships in text-rich images, we combine the text-rich image, OCR results with bounding boxes and manual caption automatically by prompting them to GPT-4o and asking for captions with more text labels while preserving the original style and structure of its manual caption. The system message used is detailed in Appendix~\ref{appendix:data_collection_details}.


Furthermore, although the manual results may contain errors, OCR bounding boxes help GPT-4o better locate text and correct inaccuracies as shown in the example of Figure~\ref{fig:pipeline}. These detail-enriched captions are further paired with descriptive questions to create single-round caption data $D^g$:$\{Q^g,A^g\}$, named \textbf{LLaVAR-2-Cap}, 
which includes detailed visual text and object descriptions.
\subsection{Visual Question Answering with Self-explanations} \label{sec:vqa}

Visual question-answering data is crucial for visual instruction tuning and typically includes extractive QA data that focuses on the local attributes of an image $I$. In LLaVAR-2-VQA, each extractive QA pair $D^e$:$\{Q^e,A^e\}$ is accompanied by a self-explain pair $D^r$:$\{Q^r,A^r\}$ to illuminate the rationale behind the extractive answer $A^e$. The detailed data collection pipeline for $D^e$ and $D^r$ is described following. 
\paragraph{Extractive Question Answering Data} \label{sec:extract}
In addition to these globally descriptive instruction-following data, instructions focusing on specific local attributes of text-rich images are also necessary. Thanks to the high-quality human-annotated captions, which clearly indicate each attribute of both text and visual areas in the image, it is convenient to utilize GPT-4o to generate conversational data that focuses on partial attributes in a local area. To achieve this, we prompt GPT-4o with the text-rich image, its human-annotated caption, OCR results from PaddleOCR, and two in-context demonstrations following \cite{zhang2023llavar}. Applying the system message in \cite{liu2023visual}, this setup generates multiple single-turn conversations $D^e$:$\{Q^e,A^e\}$ around an image, each focusing on different components without overlapping. For each single-turn conversation, we formulate the question and answer as the input instruction and the target response, respectively.
\paragraph{Self-explain Conversational Data} \label{sec:self-explain}
While the extractive instruction-following data $D^e$ focusing on local attributes will facilitate the understanding of text-rich images, it still presents challenges for MLLMs to answer extractive questions when the model needs to link a visual area with a text area or when the question involves one text area but the answer is found in another.

In these cases, precise associative and implicit reasoning is necessary. Although $D^e$ provides accurate retrieval labels, it lacks clarity on how its extraction is achieved. The answering process behind $D^e$ can supplement the extract result in $A^e$ with text/geometric labels, such as clear indications of where the query object is located in the image, thus improving MLLMs for better recognition and extraction.

To help MLLMs handle challenging scenarios, we make the implicit extraction process explicit by generating a self-explain conversation $D^r$:$\{Q^r,A^r\}$ as the explanation for each pair $D^e$. Prompting GPT-4o with the text-rich image, manual caption, OCR results, corresponding $D^e$, and system messages, the self-explain dialogue $D^r$ is generated to explain how the extractive answer $D^e$ is obtained, 
highlighting implicit connections between different components. Please refer to Figure~\ref{fig:pipeline} for an example of the two dialogues.
The self-explain dialogue $D^r$ that is based on $D^e$ will enhance MLLMs' ability to recognize implicit connections in images, improving performance in tasks such as extractive QA and text-rich image analysis, while avoiding verbosity if directly adding reasoning into extractive answers. The system message used for generating $D^r$ is detailed in Appendix~\ref{appendix:data_collection_details}.

\begin{figure*}[htp]
    \centering
    \includegraphics[width=\textwidth]{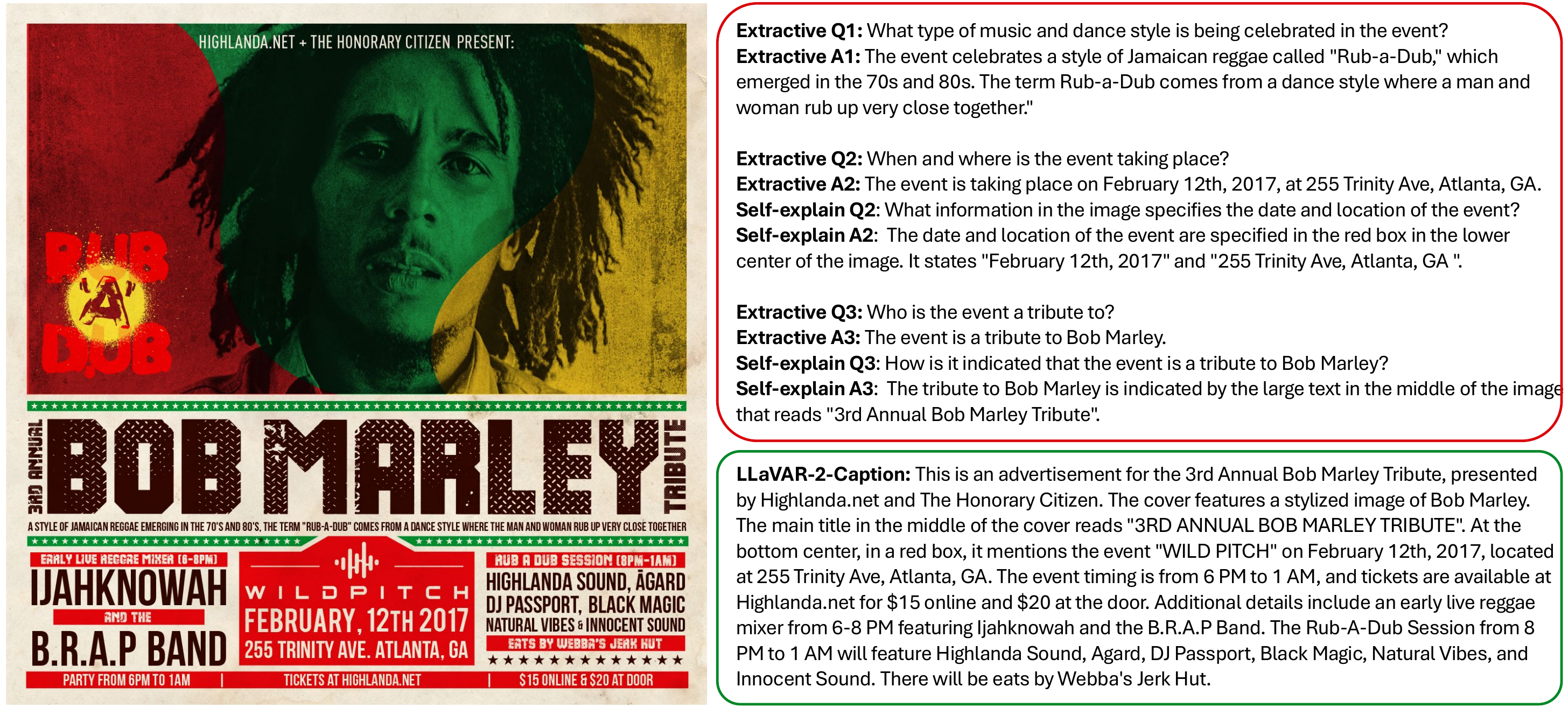}
    \vspace{-1.6em}
    \caption{A poster example of LLaVAR-2 dataset.}
    \vspace{-1em}
    \label{fig:example2}
\end{figure*}
\subsection{Instruction Data Filtering} \label{sec:explain_filter}
Both initial extractive QA $D^e$ and self-explain QA $D^r$ of LLaVAR-2-VQA generated in Section~\ref{sec:vqa} may include low-quality samples. Specifically, the low-quality extractive QA $D^e$ generally fails in two cases: (1) the entire $D^e$ is barely related to the input image $i$ but focuses on an unfamiliar or too abstract topic; (2) the extractive answer $A^e$ is anti-intuitive or unrelated to the extractive question $A^e$. While both cases may happen together, we develop an effective filtering strategy based on Instruction Following Difficulty(IFD) \cite{li2023quantity}. Given the instruction, the corresponding answer, and the model $\theta$, IFD is calculated as:
\begin{equation}
    \mathrm{IFD}_\theta(Q, A)=\frac{s_\theta(A \mid Q)}{s_\theta(A)},
\end{equation}
where $s_\theta$ is the sum of the next token prediction loss of the input computing by the model $\theta$. $s_\theta(A \mid Q)$ inputs model $\theta$ with $\{Q,A\}$ and sums the loss of each $A$'s token. $s_\theta(A)$ is computed without the instruction $A$ input. IFD can measure how helpful the instruction is towards the generation of the answer, and smaller IFD reflects better correspondence and compatibility between them. Inspired by IFD, we propose \textbf{M}ultimodal \textbf{I}nstruction-\textbf{F}ollowing \textbf{D}ifficulty (mIFD) to filter $D^e$ as: 
\begin{equation}
\begin{split}
&\mathrm{mIFD}_\theta(Q^e_i, A^e_i, I_i) \\
&= \frac{s_\theta(Q^e_i, A^e_i \mid I_i)}{s_\theta(Q^e_i, A^e_i)} \times \frac{s_\theta(A^e_i \mid Q^e_i)}{s_\theta(A^e_i)} \\
&= \mathrm{VFD}(D^e_i,I_i) \times \mathrm{IFD}(Q^e_i,A^e_i)
\end{split}
\end{equation}
In mIFD score, the first term Visual-Following Difficulty (VFD) takes the entire extractive dialogue $D^e$ and the image as the input to evaluate the correspondence between them. The second term directly utilizes the IFD score to detect samples in case (2). Computing mIFD scores on $D^e$, we excluded the highest 70$\%$ of samples. Examples filtered out are shown in Figure~\ref{fig:mifd_example}. We can find that many unrelated answers $A^e$ are generated based on MLLMs' pre-knowledge of the object.
\begin{figure}[t!]
    \centering
    \hspace{-4mm}
    \includegraphics[width=0.5\textwidth]{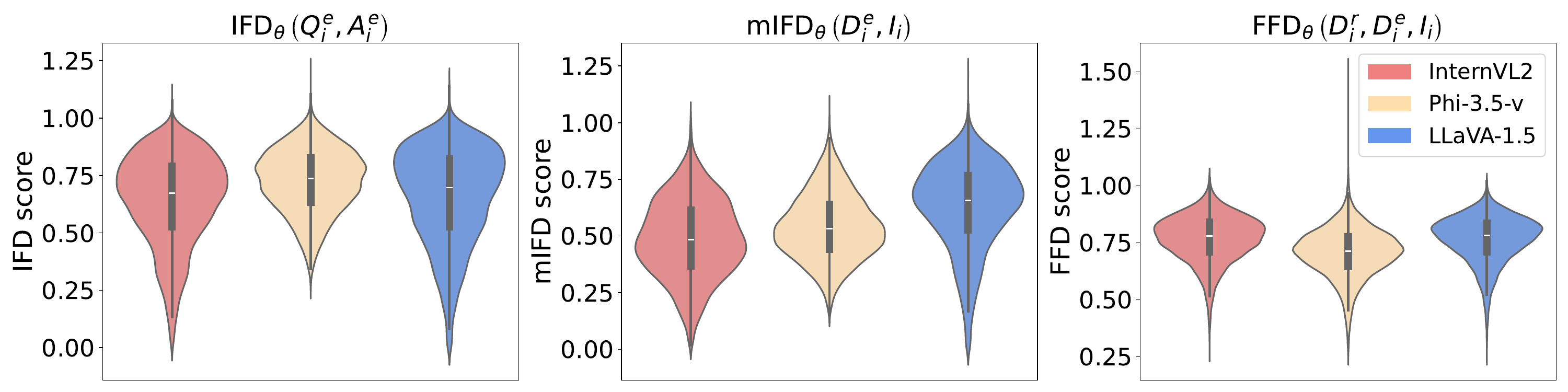}
    \caption{The distributions of $\mathrm{IFD}(Q^e_i,A^e_i)$, $\mathrm{mIFD}_\theta(Q^e_i, A^e_i, I_i)$ and $\mathrm{FFD}_\theta(D^e_i, D^r_i, I_i)$ computing by different MLLMs.}
    \vspace{-1em}
    \label{fig:filter_distri}
\end{figure}

\begin{figure}[t!]
    \centering
    \hspace{-4mm}
    \includegraphics[width=0.5\textwidth]{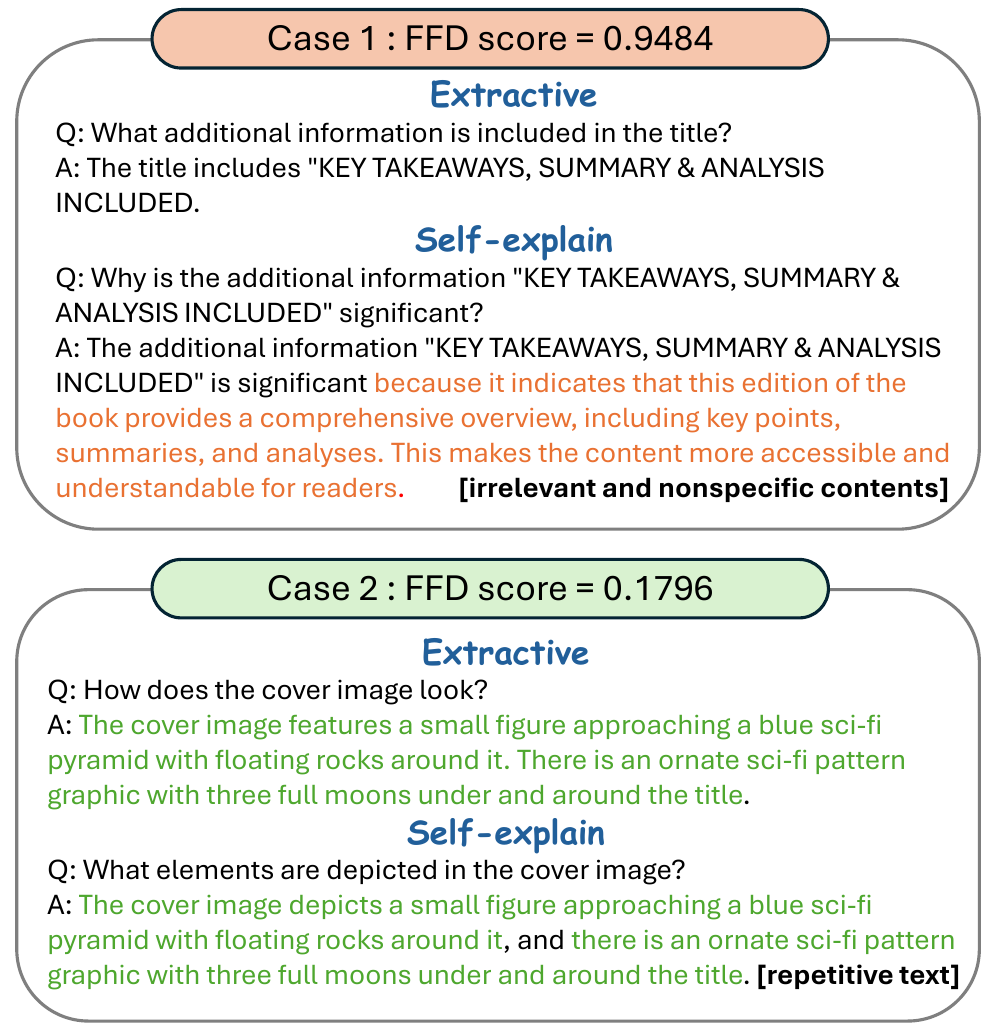}
    \caption{Example of FFD-based filtering. In case 1, the self-explain pair is composed by not specific contents barely related to the extractive pair and image; In case 2, the self-explain pair repeats the similar question-answering content. }
    \vspace{-1em}
    \label{fig:efd_filter}
\end{figure}

As for the data filtering of self-explain data $D^r$, we find abnormal $D^r$ sometimes falls into two cases: (1) $D^r$ focuses on a new concept unrelated to the image and $D^e$; (2) $D^r$ is over-corresponded with $D^e$ leading to $A^r$ very similar with $A^e$. Examples of 2 cases are shown in Figure~\ref{fig:efd_filter}. 

Extending IFD to the dialogue level, we propose the Fact-Following Difficulty (FFD) score to reflect the closeness between the extractive data $D^e$ and the self-explain data $D^r$, which is formed as:
\begin{equation}
\mathrm{FFD}_\theta(D^e_i, D^r_i, I_i)=\frac{s_\theta(D^r_i \mid D^e_i,I_i)}{s_\theta(D^r_i \mid I_i)}.
\end{equation}

From Figure~\ref{fig:efd_filter}, high FFD reflects unrelated $D^r$ and $D^e$, while low FFD score indicates repetitive answers. Utilizing Phi-3.5-vision to compute FFD, we filter out 1.5k poorly correlated and 5.6k over-related self-explain pairs $D^r$. Examples of filtered-out data are shown in Appendix~\ref{sec:appendix_filter}.

Furthermore, we verify the consistency of mIFD and FFD score computing by different MLLMs. Their distributions are shown in Figure~\ref{fig:filter_distri}. We can observe that their FFD scores share a very similar FFD pattern. While their IFD scores have slight differences, their mIFD scores computed using the IFD term keep these differences consistent, e.g. IFD computed by LLaVA-1.5 shows a narrower lower tail and this difference remains in mIFD. 
\section{Enabling LLaVA to better Read}
To verify the benefits of LLaVAR-2 for visual text understanding,  we propose \modelname{}-3.8B following the similar architecture of LLaVA~\cite{liu2023visual}. \modelname{}-3.8B is an efficient multimodal language model leveraging the small-scale but effective \texttt{microsoft/Phi-3-mini}~\cite{abdin2024phi} as the language decoder $D$ and adapting the Mixture-of-Resolution Adaptation (MRA) in LLaVA-HR~\cite{luo2024feast} to build the vision encoder $V$. Specifically, \texttt{CLIP-ViT-L} and \texttt{CLIP-ConvNeXt-L} are utilized to encode low- and high-resolution images. High-resolution features are embedded into low-resolution paths by applying MRA. Integrating high-resolution embeddings in this manner enhances the MLLM's image comprehension ability, especially for text-rich images, while maintaining efficiency without extra visual tokens. The grid embeddings before the last layer of the transformer are aligned to the word embedding space via a two-layer MLP projector. 

We conduct two-stage training following LLaVA-HR that optimizes the projector only without the MRA module in stage 1 and fully optimizes the entire MLLM during stage 2. Besides the 158K instruction-following data from LLaVA, our 40k LLaVAR-2-Cap data and 182k LLaVAR-2-VQA data are utilized in stage 1 and stage 2 training and will benefit the image text understanding capacity of \modelname{}-3.8B.

\begin{figure}[tp]
    \centering
    \vspace{-0.2em}
    \hspace{-4mm}
    \includegraphics[width=0.5\textwidth]{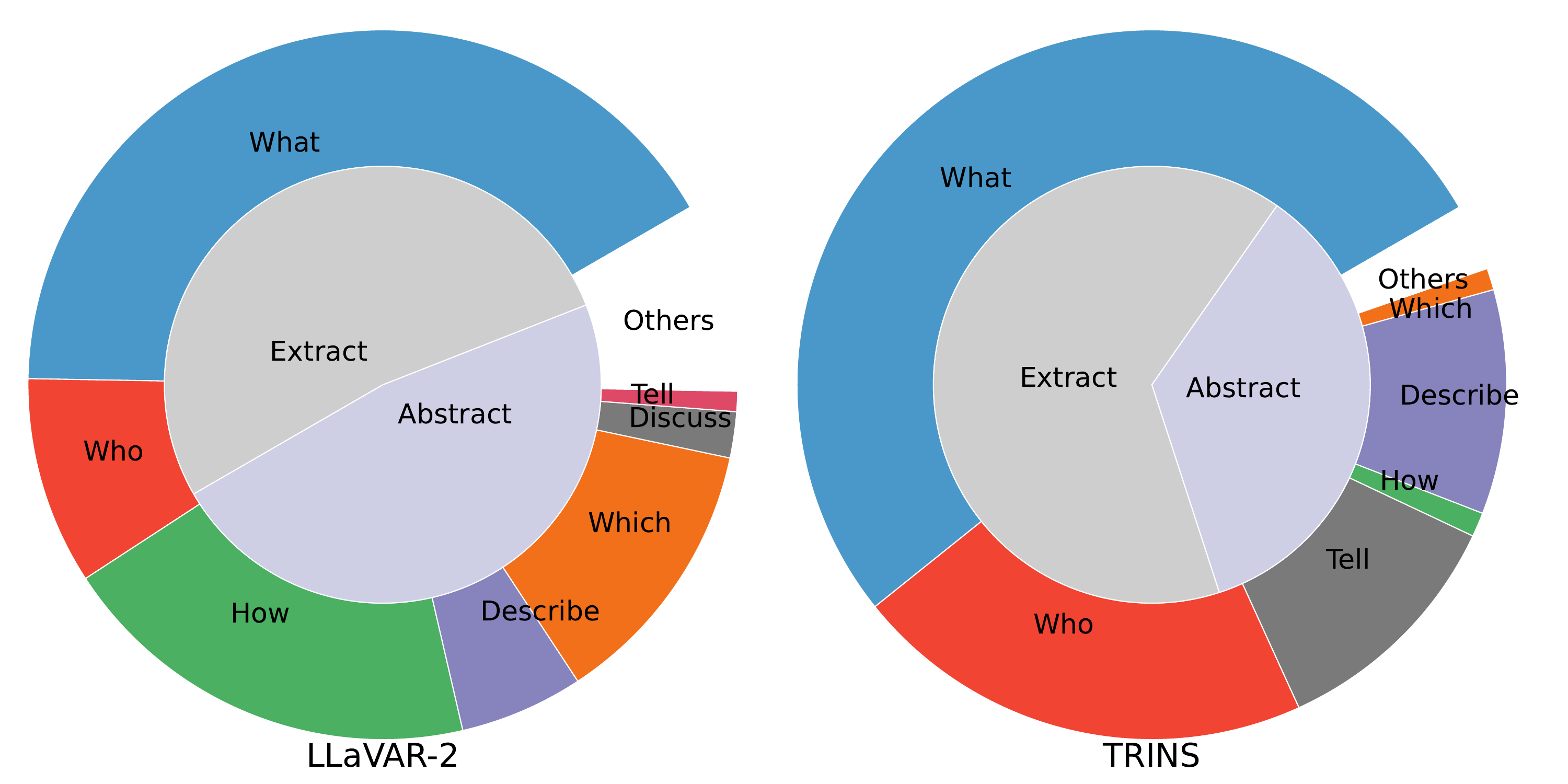}
    \caption{Instruction type statistics based on questioning words and keywords.}
    \vspace{-1em}
    \label{fig:qa_type}
\end{figure}

\section{Dataset Analysis} \label{sec:diversity_eval} 
Besides scaling dataset size, improving data diversity is essential for enhancing end-to-end visual instruction tuning in MLLMs. The variety of instructions across concepts and tasks determines a model's ability to understand and navigate text-rich images with complex semantics. In this section, we evaluate the instruction diversity of LLaVAR-2-VQA to assess its quality.

\begin{figure}[t]
    \centering
    \begin{subfigure}{0.48\columnwidth}
        \includegraphics[width=\textwidth]{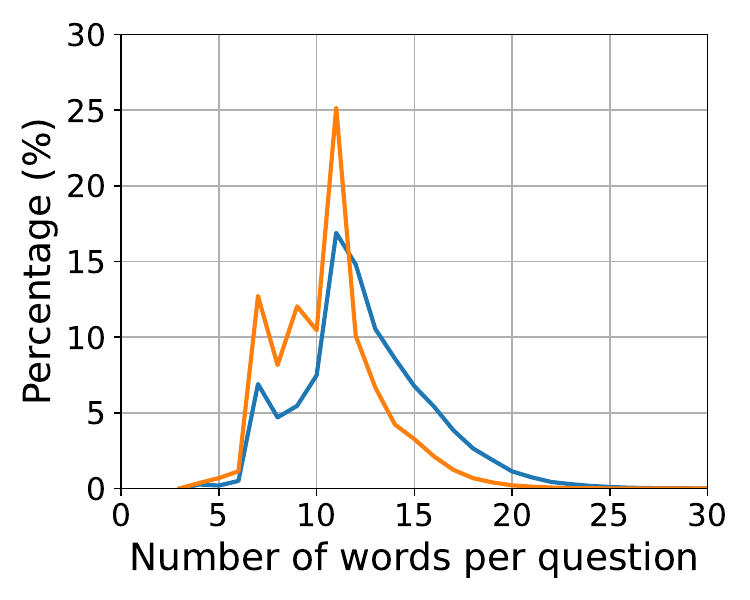}
        \caption{Question statistics} 
        \label{fig:stats3}
    \end{subfigure}
    \hfill  
    \begin{subfigure}{0.48\columnwidth}
        \includegraphics[width=\textwidth]{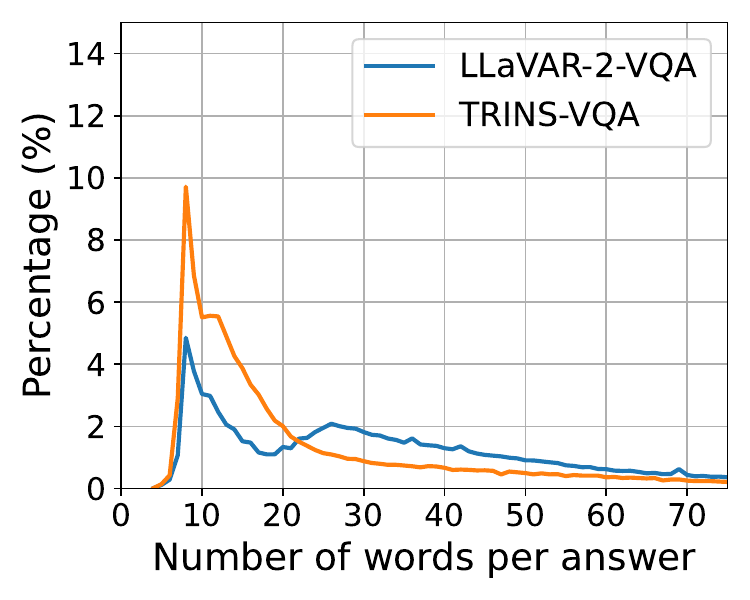}
        \caption{Answer statistics} 
        \label{fig:stats4}
    \end{subfigure}
    \vspace{-0.8em}
    \caption{Statistics for LLaVAR-2-VQA: Questions and Answers' lengths.}
    \label{fig:qa-len-stats}
\end{figure}

\paragraph{Statistics and Analysis}\label{diversity_keyword}
In Figure~\ref{fig:qa_type}, we show the visualization of instruction distribution of LLaVAR-2-VQA and TRINS-VQA~\cite{zhang2024trins}, a high-quality text visual instruction-following dataset, based on questioning words and keywords statistics following \cite{wang2022self}. The inner circle illustrates the Abstract and Extract instruction's distribution decided by keywords in questions. The outer circle reflects the distribution of questioning words in \textbf{LLaVAR-2-VQA}. It shows overall more diverse patterns of \textbf{LLaVAR-2-VQA} than \textbf{TRINS-VQA} and the emergence of self-explain data prominently enriches the diversity, e.g. self-explain questions beginning with "Which" and "How" ask for the exact image sources for the extractive answer $Q_e$. In Figure~\ref{fig:qa-len-stats}, LLaVAR-2-VQA (averages 12.4 words per question, 38.9 words per answer) shows a more balanced question and answer length distribution than TRINS-VQA (averages 10.6 words per question, 24.3 words per answer), indicating a greater variety of complexity levels in its QAs. Additional statistics are shown in Appendix~\ref{appendix:LLaVAR-2_statistcs}.
\begin{table}
    \centering
    \begin{tabular}{ccc}
        \hline Embedding Cat. & Task2Vec$\uparrow$  & S-BERT$\uparrow$ \\\hline
         Ours& 0.1444 & 0.6334 \\
         TRINS& 0.1156 & 0.5410\\\hline
    \end{tabular}
    \vspace{-0.6em}
    \caption{Instruction Diversity Coefficient of \textbf{LLaVAR-2-VQA} and TRINS-VQA based on Task2Vec and Sentence-BERT}
    \vspace{-1.2em}
    \label{tab:diversity_embedding}
\end{table}
\begin{table*}[htbp]
\small
\setlength{\tabcolsep}{4pt}
\centering
\begin{tabular}{l cc  ccccccc}
\toprule
\textbf{Method} & \textbf{Extract Acc.} & \textbf{B@1} & \textbf{B@2} & \textbf{B@3} &\textbf{ B@4} & \textbf{METEOR} & \textbf{ROUGE} & \textbf{CIDEr} \\
\hline
Phi-3-vision-128k \cite{abdin2024phi} &22.3 &47.2  &35.8  & 29.1 & 24.4 & 51.5 &63.5  &293.8 \\
Phi-3.5-vision \cite{abdin2024phi} &18.7&42.6  &32.1  & 26.0 & 21.7 & 47.7 &61.1  &280.8 \\
LLaVA-1.5-7B \cite{liu2023visual} &7.8 & 34.5 &23.9  & 17.9 &14.1 &44.0 &46.1&123.0 \\
LLaVA-NeXT-7B \cite{liu2024llavanext} &9.3 & 36.9 &26.5  & 20.6 &16.7 &49.4 &50.4&142.2 \\
Idefics3-8B \cite{laurenccon2024building}&32.1 & 28.2&22.4  & 18.9 &16.4 &44.5 &55.1&222.4\\
InternVL2-8B \cite{chen2024far}&25.3& 31.1  &22.3  &17.2  &13.9& 55.1&52.1&174.6 \\
MiniCPM-V-2.6-8B \cite{yao2024minicpm}&42.7& 40.4 & 30.3 & 24.5 &20.6 &61.0 &60.4&259.6 \\
Cambrian-1-8B \cite{tong2024cambrian}&43.8& 40.4  &30.5  & 24.7 &20.8&56.4 &57.7& 222.3\\
\hline
\modelname{ (Llama-3.1 8B)} $\ddagger$&59.0&53.6&43.1&36.6&31.9&64.5&68.2&355.5&\\
\modelname{ (Vicuna-1.5 13B)}$\ddagger$&\textbf{62.1}&\underline{55.2}&\textbf{45.0}&\textbf{38.4}&\textbf{33.7}&\underline{65.2}&\textbf{69.6}&\textbf{371.0}&\\
\rowcolor{gray!30} 
\modelname{ (Phi-3-Mini) w/o $D^{r}$} $\ddagger$ &53.2  &52.5  &42.1  & 35.5 & 30.8 & 64.8 &67.9  & 349.1 \\
\modelname{ (Phi-3-Mini)} $\ddagger$ &\underline{59.6}  &\textbf{55.4}  &\underline{44.8}  & \underline{38.1} & \underline{33.3} & \textbf{65.4} & \underline{69.1} &  \underline{362.5}\\
\bottomrule
\end{tabular}
\vspace{-0.7em}
\caption{Results of \modelname{} models trained w/wo self-explain data $D^{r}$ for text-rich image question-answering tasks. We use $\ddagger$ to refer to models fine-tuned on the proposed LLaVAR-2 dataset. We use \textbf{bold} and \underline{underline} to indicate the best and second best results, respectively. Row in the gray is the ablation result without $D^{r}$.}
\label{tab:LLaVAR-2-VQA}
\end{table*}

\begin{table*}[ht]
\small
\centering
\begingroup
\setlength{\tabcolsep}{4pt} 
\renewcommand{\arraystretch}{1} 
\begin{tabular}{l cccccccc}
\toprule 
                   &  \textbf{ST-VQA}  & \textbf{TextVQA} & \textbf{DocVQA} & \textbf{ChartQA} & \textbf{InfoVQA} & \textbf{FUNSD} & \textbf{SROIE}\\ \midrule
BLIP-2 \citep{li2023blip} $\dagger$                  & 21.7                        & 32.2             & 4.9        & 3.4 & 11.3 & 0.20 & 0.14     \\
OpenFlamingo \citep{awadalla2023openflamingo} $\dagger$                      & 19.3                         & 29.1             & 5.1   & 9.1 & 15.0 & 0.85 & 0.12          \\
MiniGPT4 \citep{zhu2023minigpt} $\dagger$                    & 14.0                           & 18.7             & 3.0  & 4.3 & 13.3 & 1.19 & 0.04             \\
mPLUG-Owl \citep{ye2023mplugowl} $\dagger$                            & 29.3                        & 40.3             & 6.9 &9.5 & 16.5 & 1.02 & 0.60            \\ 
LLaVA \citep{liu2023visual} $\dagger$                 & 28.9                           & 36.7             & 6.9 &    28.9 & 13.8 & 1.02 & 0.12\\
LLaVA1.5 \citep{liu2023visual} $\dagger$                 & 38.1                           & 38.7             & 8.5   & 9.3 & 14.7 & 0.20 & 1.70  \\
LLaVAR \citep{zhang2023llavar}$\dagger$               & 39.2  & 48.5 & 11.6 & 12.2 & 16.5 & 0.50 & 5.20    \\
mPLUG-Owl2 \citep{ye2023mplug} $\dagger$                            & 29.3                         & 40.3             & 6.9  & 19.4 & 18.9 & 1.40 & 3.20                  \\
Monkey~\cite{li2024monkey}$\dagger$ & 54.7  & 64.4 & 50.1 & 54.0 & 25.8 & 24.1 & 41.9 \\
TextMonkey~\cite{liu2024textmonkey} $\dagger$& \textbf{61.8}& \textbf{71.3}& 64.3 &58.2& 28.2  & 32.3 & 47.0  \\
LaRA-13B \cite{zhang2024trins} $\dagger$ &47.2& 59.9& 50.8& 25.6&28.4& 23.2&36.6\\
\midrule
\modelname{ (Llama-3.1 8B)}&51.6&61.0 &\underline{69.3}&69.9&\underline{32.0}&32.3&\underline{58.4}\\
\modelname{ (Phi-3-Mini)} & 52.3 & 60.2 & 66.1 & \textbf{78.5} & 30.3 & \underline{36.0} & 51.9 \\
\modelname{ (Vicuna-1.5 13B)}&\underline{59.5}& \underline{66.0}& \textbf{71.3} & \underline{76.3} & \textbf{40.2} & \textbf{36.7} & \textbf{61.9}&\\
\bottomrule
\end{tabular}
\endgroup
\vspace{-0.7em}
\caption{\label{table: VQA result} Zero-shot performance (accuracy \%) on text-based VQA. We use $\dagger$ to refer to the results obtained from previous work~\cite{liu2023hidden}, and use \textbf{bold} and \underline{underline} to indicate the best and second best results, respectively.}
\label{tab:vqa_result}
\vspace{-1.0em}
\end{table*}
\normalsize
\paragraph{Diversity Evaluation via Embedding Distance Measurement}\label{diversity_embedding}
 The statistics and visualization in Section~\ref{diversity_keyword} provide clear sights of how the \textbf{LLaVAR-2-VQA} is diversely composed. However, this evaluation based on single-word extraction using manual or neural parsers is not representative enough to reflect the precise instruction coverage. To study a more accurate diversity evaluation of \textbf{LLaVAR-2-VQA}, we use the intra-dataset diversity coefficient~\cite{lee2023beyond} to compute the dataset diversity indicator based on the task-specific TASK2VEC~\cite{achille2019task2vec} embedding and the general sentence-level Sentence-BERT~\cite{reimers2019sentence} embedding. Specifically, we formed the intra-dataset instruction diversity coefficient following~\cite{lee2023beyond} as: 
\begin{equation}
    \operatorname{div}(D)=\mathbb{E}_{B_1, B_2 \sim D} d\left(f_{B_1}, f_{B_2}\right),
    \label{eq:diversity_coefficient}
\end{equation}
 where $B_1$ and $B_2$ are batches sampled from the target dataset $D$ and $f_{B_i}$ denotes the batch-level embedding which is the diagonal of Fisher Information Matrix for TASK2VEC or the mean of the instructions' Sentence-BERT embeddings of the batch. We apply Cosine distance  $d$ to measure the semantic gap between sampled batches within the target dataset. Thus, larger $ \operatorname{div}(D)$ indicates higher instruction diversity. While TASK2VEC provides fine-grained task-level diversity measurement, Sentence-BERT indicates general semantic diversity. 

 We compare the instruction diversity coefficient \textbf{LLaVAR-2-VQA} with \textbf{TRINS-VQA} in Table~\ref{tab:diversity_embedding}, with batch size $\|B\|=512$ and 200 batches for each dataset. As shown, \textbf{LLaVAR-2-VQA} exhibits a higher diversity level under both embedding settings, indicating the enriched instruction category and semantics discussed in Section~\ref{diversity_keyword}.
\section{Experimental Results}
We aim to illustrate the benefits of integrating LLaVAR-2 in visual instruction tuning. \modelname{} models are evaluated on LLaVAR-2-Cap, LLaVAR-2-VQA and classical text-rich benchmarks to demonstrate the effect of LLaVAR-2 to MLLMs. All experiments are implemented with PyTorch and performed on Nvidia A100 GPUs.

\subsection{Visual Question Answering}
We first evaluate \modelname{} models and baselines \modelname{} in LLaVAR-2-VQA shown in Table~\ref{tab:LLaVAR-2-VQA}. We report Extract Accuracy following \cite{wu2023ocr}, text similarity metrics BLEU \cite{papineni2002bleu}, METEOR \cite{banerjee2005meteor}, ROUGE \cite{lin2004rouge}, and CIDEr \cite{vedantam2015cider}. \modelname{} with Phi-3-Mini as the backbone outperforms other methods in zero-shot VQA on LLaVAR-2-VQA. 
Thanks to the enriched details and the implicit answer process in LLaVAR-2, \modelname{} models can have a better comprehension of visual texts and generate precise extractive answers reflecting from the metric of Extract Accuracy. 
While sentence similarity metric's results can reflect how well the model can deal with complex extract questions and to what extent the model understands the reasoning process behind the extractive result regarding the self-explain QA set, \modelname{}-3.8B's better results on these metrics demonstrate the benefits of LLaVAR-2 on these two perspectives and also indicates the interplay of extractive and self-explain pairs in LLaVAR-2-VQA. Comparing the performance of \modelname{} and \modelname{} w/o $D^{r}$, the supplement of the self-explain pair $D^r$ after $D^e$ can enhance the model's capacity toward text-rich images. As for other methods, MiniCPM-V \cite{yao2024minicpm}, Cambrian-1 \cite{tong2024cambrian} , and Phi-3/3.5-vision \cite{abdin2024phi} perform well on LLaVAR-2-VQA, highlighting the importance of including diverse visual text data during training.
\begin{table}[ht]
\centering
  \setlength{\tabcolsep}{2pt}
  \small
  \resizebox{0.48\textwidth}{!}{
  \begin{tabular}{ccccccc}
\hline       Method
             & Recog. & VQA$^{S}$ & VQA$^{D}$ & KIE & Total \\ \hline
Gemini       & 215                   & 174                         & 128                   & 134               & 651               \\ 
GPT-4v        & 167                   & 163                         & 146                   & 160              & 636               \\
\hline
Text-Monkey       & 169                   & \textbf{164}                         & \underline{115}                    & 116              & 561               \\
Monkey       & 174                   & {161}                         & {91}                    & 88              & 514               \\
mPLUG-Owl2   & 153                   & 153                         & 41                    & 19                & 366               \\
LLaVAR      & 186                   & 122                         & 25                    & 13               & 346               \\
LLaVA1.5-13B & 176                   & 129                         & 19                    & 7                & 331               \\
MiniGPT-V2   & 124                   & 29                          & 4                     & 0                & 157               \\
LaRA-13B   & {206}                   & 151                         & 101                    & \underline{145}               & {603}               \\
\hline
\modelname{} (Phi-3-Mini) & \underline{225}                   & {152}                         & {114}                    & {143}               & \underline{634} \\
\modelname{} (Vicuna-1.5) & \textbf{241}                   & \underline{162}                         & \textbf{121}                    & \textbf{156}               & \textbf{680}   
\\
\hline
\vspace{-1.5em}
\end{tabular}}
\captionof{table}{Results of MLLMs on OCRBench. Recog. represents text recognition, VQA$^{S}$ represents Scene Text-Centric VQA, VQA$^{D}$ represents Document-Oriented VQA. We use \textbf{bold} and \underline{underline} to indicate the best and second best results, respectively.}
\label{tab:OCRBench}
\end{table}

\begin{table*}[htbp]
\small
\setlength{\tabcolsep}{7pt}
\centering
\begin{tabular}{l c ccccccc}
\toprule
  \textbf{Method}& \textbf{B@1} & \textbf{B@2} & \textbf{B@3} &\textbf{ B@4} & \textbf{METEOR} & \textbf{ROUGE} & \textbf{CIDEr} \\
\hline
Phi-3-vision-128k &43.6  &27.7  & 18.5 & 13.0 & 34.3 &51.6  & 35.9\\
Phi-3.5-vision &40.5  &25.5  & 16.8 & 11.7 & 32.2 &49.8  & 30.9\\
LLaVA-1.5-7B & 28.3 &14.7  & 7.9 &4.9 &21.8 &36.6&10.2 \\
LLaVA-NeXT-7B & 30.8 &19.8  & 13.6 &10.0 &28.1 &45.0&27.7 \\
Idefics3-8B &  21.8 &15.4  & 11.0 &8.2 &36.2 &31.6&1.4\\
InternVL2-8B& 21.9 & 14.8 &10.2  &7.5 &37.7 &35.9&4.9 \\
MiniCPM-V-2.6-8B& 41.0 & 27.6 & 20.1 &15.4 &34.8 &48.9& 28.9\\
Cambrian-1-8B&36.6& 23.2  &15.6  & 11.1 &31.0&47.8 &28.8 \\
\hline
\modelname{ (Llama-3.1 8B)} $\ddagger$&\underline{58.7}&\underline{43.6}&\underline{33.8}&27.1&47.1&60.5&\underline{61.4}&\\
\modelname{ (Vicuna-1.5 13B)}$\ddagger$&\textbf{58.9}&\textbf{44.4}&\textbf{34.9}&\textbf{28.3}&\textbf{48.0}&\textbf{61.7}&\textbf{66.9}&\\
\rowcolor{gray!30}
\modelname{ (Phi-3-Mini)} w/o $D^{g}$ $\ddagger$&31.6  &20.1  & 14.2 & 10.4 & 32.1 &52.2  &39.7 \\
\modelname{ (Phi-3-Mini)} $\ddagger$&58.1  &43.4  & \underline{33.8} & \underline{27.2} & \underline{47.2} & \underline{60.9} & 60.8\\
\bottomrule
\end{tabular}
\vspace{-0.7em}
\caption{Results of \modelname{} models trained w/wo global instruction-following data $D^{g}$ based on detailed captions for text-rich image captioning tasks. We use $\ddagger$ to refer to models fine-tuned on LLaVAR-2 dataset, use \textbf{bold} and \underline{underline} to indicate the best and second best results, respectively. Row in the gray is the ablation result without $D^{g}$.}
\vspace{-1.8em}
\label{tab:LLaVAR-2-cap}
\end{table*}

In addition, we also conduct evaluations on the classical visual text understanding benchmarks \cite{liu2023visual} and OCR-Bench \cite{liu2023hidden} shown in Table~\ref{table: VQA result} and Table~\ref{tab:OCRBench}. 
For benchmarks involving images that include both textual and abstract visual elements, such as DocVQA \cite{mathew2021docvqa}, ChartQA \cite{masry2022chartqa}, and InfoVQA \cite{mathew2022infographicvqa}, \modelname{} models significantly outperform other methods, highlighting the importance of high proportion of abstract QAs within LLaVAR-2-VQA to improve abstract document understanding. Additionally, while these three datasets are involved in the training of Monkey \cite{li2024monkey} and TextMonkey \cite{liu2024textmonkey}, \modelname{} performs better than these finetuned models on these three datasets, thanks to high-quality text-centric VQAs in LLaVAR-2-VQA/Cap.
For scene text-centric ST-VQA \cite{biten2019scene} and TextVQA \cite{singh2019towards} which are dominated by natural scene/object images with text, \modelname{} models have the comparable performance with Monkey and TextMonkey, which includes TextVQA during their training. 
Surprisingly, we find \modelname{} model is good at reading on scanned pure text images even noisy ones, such as those in FUNSD \cite{jaume2019funsd} and SROIE \cite{huang2019icdar2019}. Combined with the result of OCRBench, LLaVAR-2-VQA shows comprehensive improvement in fine-tuning for Visual Question Answering of text-rich images. Furthermore, Monkey and TextMonkey show comparable results on benchmarks such as ST-VQA and TextVQA, due to the rich text labels involved in their multi-task training. LLaVAR-2 shares similar ideas: the enriched detail in LLaVAR-2-Cap and the self-explain answers in VQA supplement considerable text labels.
\subsection{Text-rich Image Captioning}
We evaluate the captioning capacity of \modelname{} models and baselines on LLaVAR-2-Cap, which requires MLLMs to generate summaries and keep necessary text labels at the same time. We compare the performance on this challenging task in Table~\ref{tab:LLaVAR-2-cap}. We can observe that MiniCPM-V, Cambrian-1, and Phi-3/3.5-vision achieved remarkable improvement upon classic MLLMs such as LLaVA-1.5. It is due to the considerable attention of these models on text-heavy visual tasks, for example, MiniCPM-V \cite{yao2024minicpm} includes a large amount of text-rich captioning data in its stage-1 and 2 training, Phi-3.5-vision's post-training is involved with diverse text image tasks. Fine-tuned on LLaVAR-2-Cap, \modelname{} models outperform other methods on all the text similarity metrics, indicating the necessity of the comprehensive captions in LLaVAR-2-Cap. Besides, the Mixture-of Resolution Adaptation used in the visual encoder of \modelname{} models adopted from LLaVA-HR plays a crucial role in making accurate summaries in captioning tasks. Among \modelname{} models with different backbones,  \modelname{} (Vicuna-1.5 13B) achieved the best result, while smaller-sized backbones such as Phi-3-Mini and Llama-3.1 8B still obtained comparable performance. \modelname{ (Phi-3-Mini)} w/o $D^{g}$ is only fine-tuned on the VQA data without adaptation on captioning data. Its not ideal result on captioning tasks further demonstrates the essential of global descriptive data for summarizing tasks.
\subsection{Additional Experiments on Data Filtering}
\begin{figure}[t]
    \centering
    \begin{subfigure}{0.48\columnwidth}
        \includegraphics[width=\textwidth]{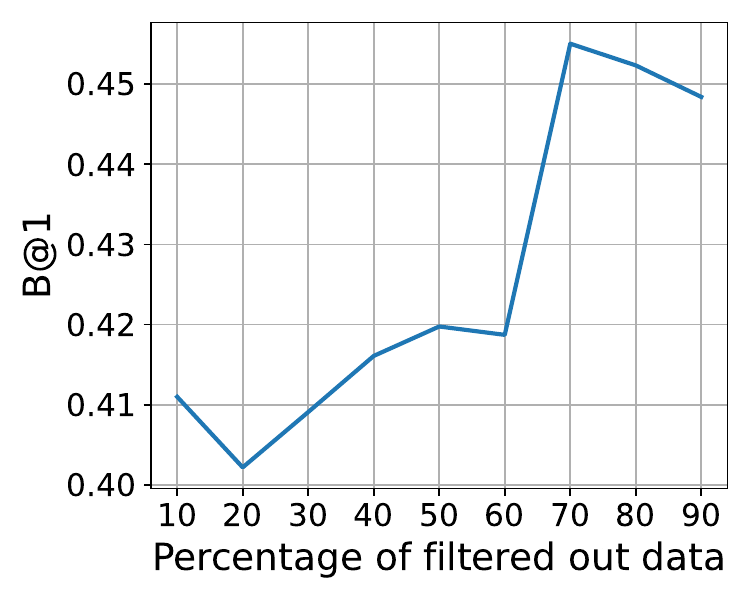}
        \vspace{-1.8em}
        \caption{BLEU-1} 
        \label{fig:bleu1}
    \end{subfigure}
    \hfill  
    \begin{subfigure}{0.48\columnwidth}
        \includegraphics[width=\textwidth]{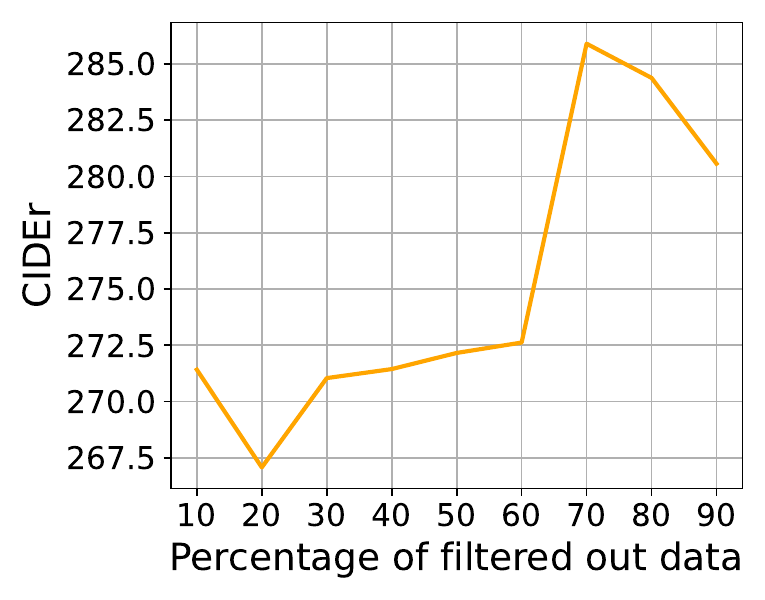}
        \vspace{-1.8em}
        \caption{CIDEr} 
        \label{fig:CIDEr}
    \end{subfigure}
    \vspace{-0.8em}
    \caption{\modelname's performance on LLaVAR-2-VQA with different percentages of filtered-out data.}
    \vspace{-1.5em}
    \label{fig:ablation}
\end{figure}

In this section, we aim to verify the effectiveness of the proposed mIFD and FFD scores. We apply the mIFD score to select the extractive data $D^e$ first and then filter $D^r$ via FFD score. Filtering out data ranging from 10$\%$ to 90$\%$, We apply the different filtered LLaVAR-2-VQA train set as the only fine-tuning dataset for different checkpoints to verify the effectiveness of the proposed filtering method. In Figure~\ref{fig:ablation}, we present the BLEU-1 and CIDEr results for different filtering percentages. It is evident that 70$\%$ is the sweet spot for LLaVAR-2-VQA that our filtering scores are efficient before 70$\%$ and after 70$\%$ performance drops due to the limited size of data for fine-tuning. These results demonstrate that the proposed filtering score can select high-quality samples enabling the model to reach better performance with a smaller data size.

\section{Conclusions}
We propose Hybrid-Instruct, an automatic multi-modal instruction generation framework based on manual captions for visual instruction tuning tailored to text-rich images, to construct LLaVAR-2 data. The detail-enriched captions and self-explain dialogues in LLaVAR-2 enhance the performance of MLLMs on different benchmarks. In addition, the proposed filtering score mIFD and FFD are shown to be effective in filtering out unqualified dialogues in the LLaVAR-2 given the image and extraction contexts. In general, LLaVAR-2 introduces novel approaches to generate and select diverse and high-quality instruction-following data for MLLMs.  
\newpage
\section{Limitations}
While the LLaVAR-2 dataset offers significant improvements in MLLMs' tuning, It still leaves us limitations to improve: (1)the quality of OCR results we relied on to augment manual captions and to create self-explain dialogues may include errors. (2) LLaVAR-2's data generation is dependent on strong MLLMs such as GPT-4o, which possibly introduces biases and is expensive to use. (3) Our proposed filtering process only occurs during post-filtering. The filtered-out samples are wasted and should still be leveraged.  

Thus, for future work, we will focus on designing self-correction data collection pipelines. The potential solutions could be adding an extra checking module to ensure the quality of inputs and utilizing the filtered data to fine-tune a small rating model for better data selection.
\bibliography{anthology,custom}
\clearpage
\appendix

\section{LLaVAR-2 Statistics}
\label{appendix:LLaVAR-2_statistcs}
\begin{table}[H]
\small
\centering
\begingroup
\setlength{\tabcolsep}{7pt} 
\renewcommand{\arraystretch}{1} 
\begin{tabular}{lcc}
\toprule
Number of images & 42870 \\
Number of detail-enriched captions & 42870 \\
Number of LLaVAR-2-Cap pairs & 42870 \\
Average \# of words per detail-enriched captions & 114.5 \\
Number of LLaVAR-2-VQA pairs & 382406  \\
Average \# of words per VQA question & 12.4\\
Average \# of words per VQA answer & 38.9\\
\bottomrule
\end{tabular}
\endgroup
\vspace{-1em}
\caption{LLaVAR-2 Dataset Statistics}
\vspace{-1em}
\label{tab: stats}
\end{table}

\section{Details for LLaVAR-2 Data Collection}
\label{appendix:data_collection_details}
\newtcolorbox{myquote}{
    breakable,
    width=0.5\textwidth,
    colback=white,
    colframe=black,
    fontupper=\itshape,
    boxrule=0.2mm,
    left=1mm,
    right=1mm,
    arc=3mm,
    auto outer arc
}
\paragraph{System message for the generation of detail-enriched captions $A^g$ in LLaVAR-2-Cap:}
\begin{center}
\begin{myquote}

You are an AI visual assistant, and you are seeing a single image. What you see is provided with an image, one corresponding OCR result, and one corresponding image caption describing the information within the same image you are looking at. Image captions and OCR results might include hallucinations, while OCR results are more accurate. OCR results are constructed with [[4 bounding-box coordinates], text, OCR confidence]. Enrich the image caption with detailed support text and location information from the OCR result.

The enriched image caption should be in a tone that a visual AI assistant is seeing the image and describing the image.  
\begingroup
\setlength{\parskip}{0pt}
Maintain the content in the original image caption while adding the support information from the OCR result to its corresponding part in the original image caption for a detailed description:

(1) for the support text added to the image caption, its bounding box coordinates should indicate the similar location described in the corresponding part of the original image caption;

(2) DO NOT mention OCR bounding box coordinates in your caption. Provide the location information with the same style as the original image caption;

(3) DO NOT add information that looks unrelated to or contradicts OCR results;

(4) When OCR results include new information, enrich this content into the original caption;

(5) When OCR results and image caption has large difference in describing the same area of the image, maintain the description in the original image caption;

(6) Do not include garbled characters in the generation.
\endgroup
\end{myquote}
\end{center}

\paragraph{System message for the generation of extractive conversation $D^e$ in LLaVAR-2-VQA following \cite{liu2023visual}:}
\begin{center}
\begin{myquote}

You are an AI visual assistant, and you are seeing a single image. What you see is provided with two OCR results and one image caption describing the information within the same image you are looking at. Image captions might include hallucinations, while OCR results are more accurate. Answer all questions with definite answers as you are seeing the image.

Design a conversation between you and a person asking about this photo. The answers should be in a tone that a visual AI assistant is seeing the image and answering the question. Ask diverse questions and give corresponding answers.

\begingroup
\setlength{\parskip}{0pt}
Include questions asking about the visual content of the image (e.g., the man, the sunset, the ocean.) and the texts contained in the image. Only include questions that have definite answers:

(1) one can see the content in the image that the question asks about and can answer confidently;

(2) one can determine confidently from the image that it is not in the image. Do not ask any questions that cannot be answered confidently;

(3) DO NOT mention OCR or image caption in your questions and answers;

(4) DO NOT ask about information from captions while it looks unrelated to or contradicts OCR results.
\endgroup

Also include complex questions that are relevant to the content in the image, for example, asking about background knowledge of the texts in the image, asking to discuss the design of the image, etc. Again, do not ask about uncertain details. Provide detailed answers when answering complex questions. For example, give detailed examples or reasoning steps to make the content more convincing and well-organized. You can include multiple paragraphs if necessary.
\end{myquote}
\end{center}

\paragraph{System message for the generation of self-explain conversation $D^r$ in LLaVAR-2-VQA:}
\begin{center}
\begin{myquote}
You are an AI visual assistant, and you are seeing a single image. What you see is provided with an image, one corresponding OCR result, one corresponding image caption describing the information within the same image you are looking at, and a set of reference QAs on this image. OCR results contain text description with location information which is constructed with [[4 bounding-box coordinates], text, OCR confidence]. The image caption contains a visual description from the human. A set of reference questions and answers around this image are provided after the image, OCR result, and image caption. Based on them, generate a pair of reasoning QA that asks why/how/where each provided reference answer looks like it and explains it in the generated reasoning answer

\begingroup
\setlength{\parskip}{0pt}
The generated reasoning QA should be in a tone that a visual AI assistant is seeing the image and answering the question:

(1) The generated question can ask the reason behind the provided reference QA's answer or ask how to get the provided QA's answer according to the provided context;

(2) If the reasoning process for the reference QA is very explicit, the generated reasoning QA should focus on the source of the reference answer, e.g. for simple extractive question answering, the reasoning QA should focus on where the extraction of the target object of the reference question from in the image and its content in the image;

(3) If the reasoning process for the reference QA is complex, the generated reasoning QA should reason on the relationship/connection between the involved objects (text object, visual object if needed) in the image;

(4) The generated questions should have diverse styles, do not only use a single style format;

(5) Find supportive information from the provided image/caption/OCR results to answer the generated reasoning question, but only assuming you can see the image, do not mention that you can see the caption/OCR results;

(6) Make sure the generated reasoning QA corresponds/very related to the reference QA and the image, avoid only extending new concepts in the reasoning QA;

(7) Generated reasoning QA should be in the same format as the provided QA:'Reasoning Question: $\{$generated reasoning question$\}$\texttt{\textbackslash\textbackslash n}
Reasoning Answer: $\{$generated reasoning answer$\}$'.
\endgroup
\end{myquote}
\end{center}
\section{LLaVAR-2-Cap Examples} \label{appendix:gd_example}
\begin{figure}
    \centering
    \includegraphics[width=1\linewidth]{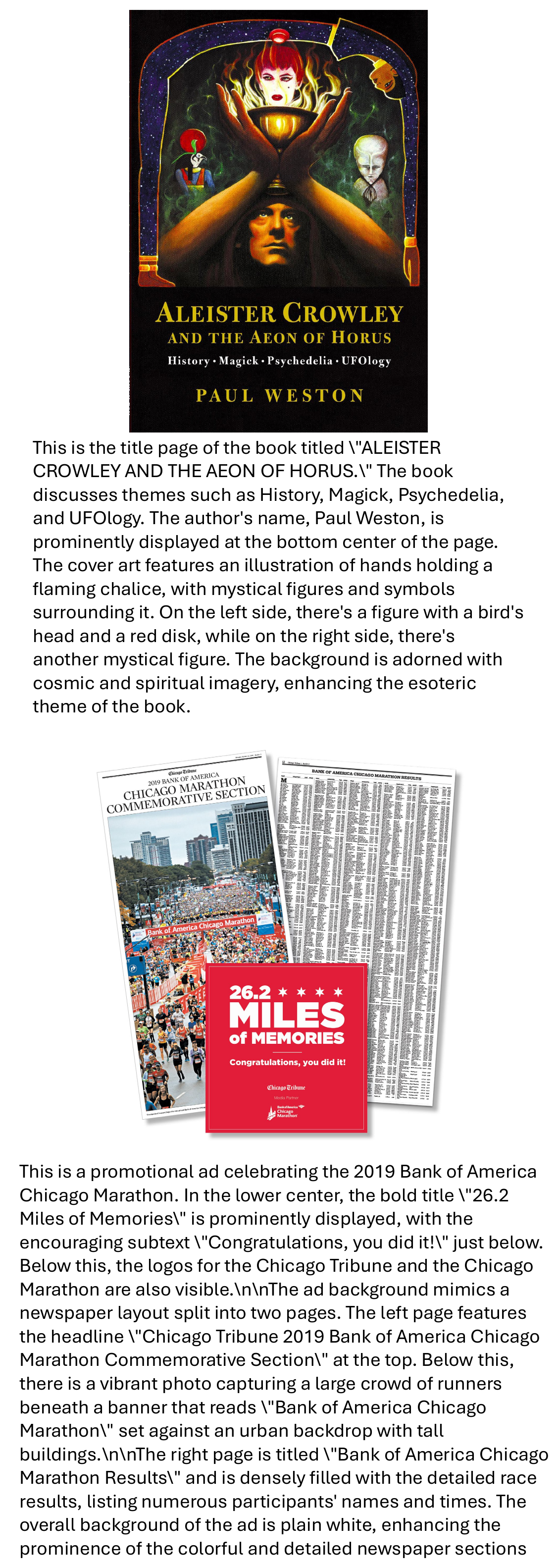}
    \caption{Caption examples in LLaVAR-2-Cap}
    \label{fig:cap1}
\end{figure}

\begin{figure*}[!ht]
    \centering
    \hspace{-4mm}
    \includegraphics[width=\textwidth]{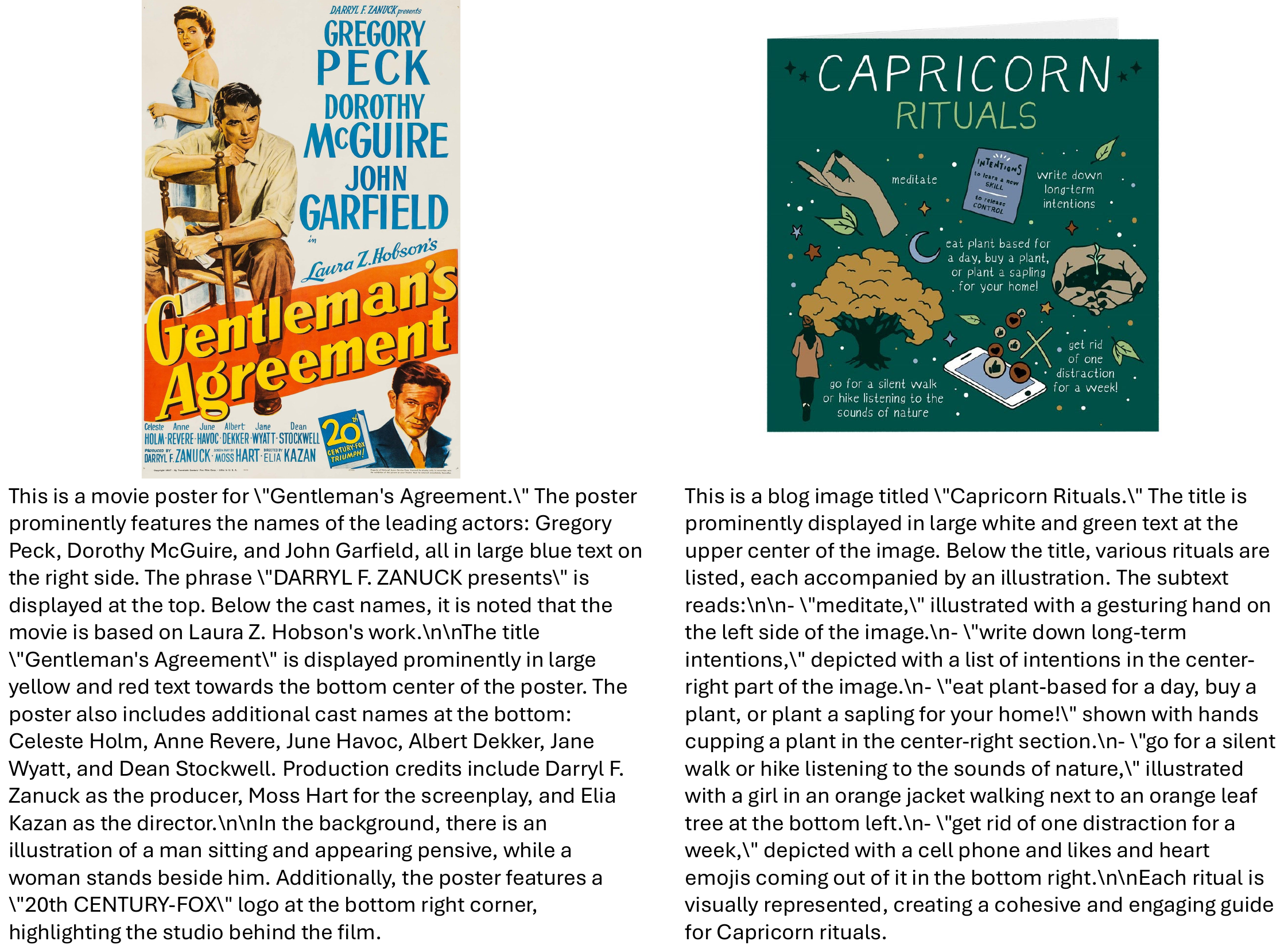}
    \hspace{3em}
    \includegraphics[width=\textwidth]{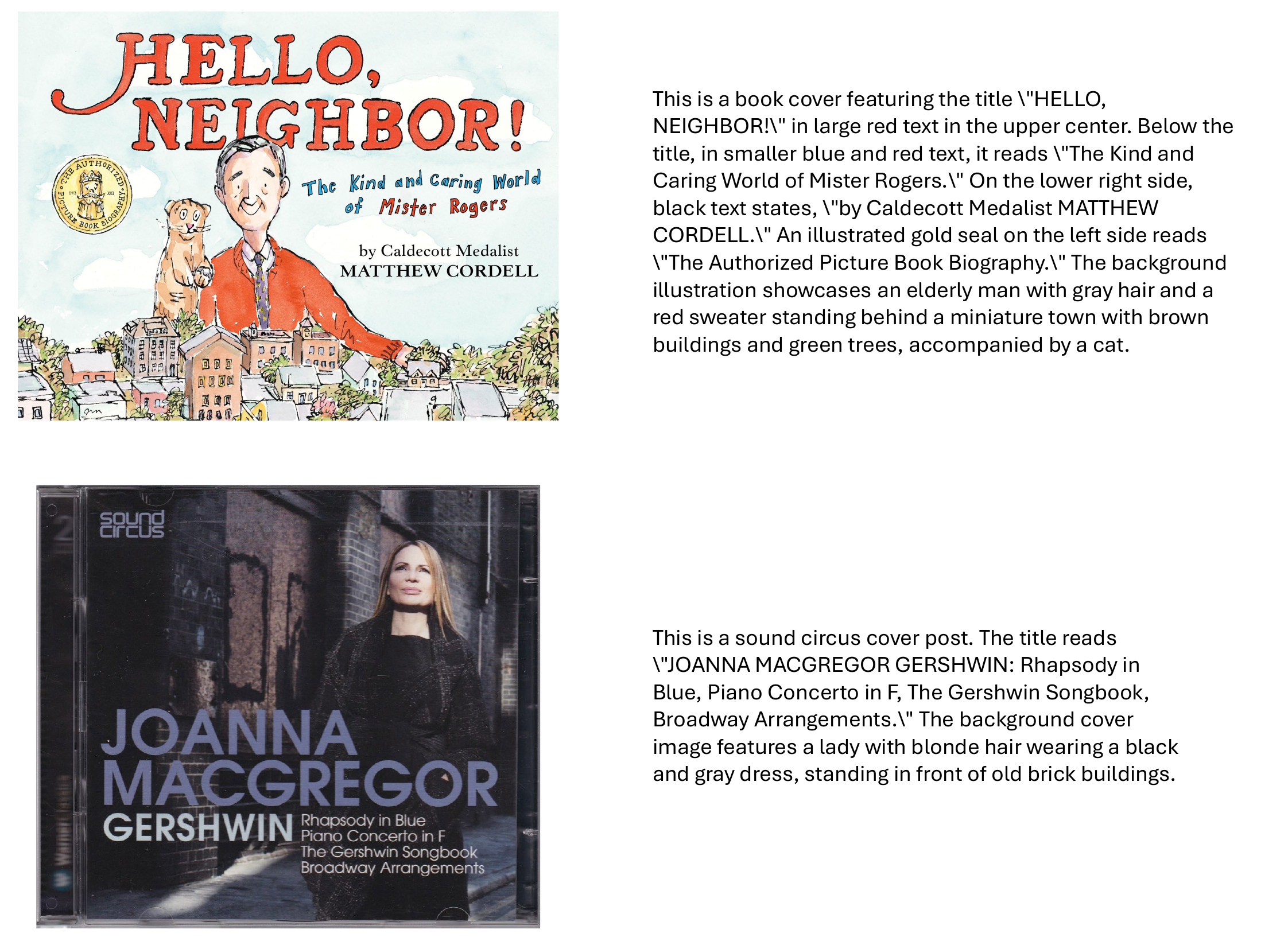}
    \caption{Caption examples in LLaVAR-2-Cap}
    \label{fig:cap2}
\end{figure*}

Figure~\ref{fig:cap1} and Figure~\ref{fig:cap2} present examples in LLaVAR-2-Cap.

\begin{figure*}[!ht]
    \centering
    \hspace{-4mm}
    \includegraphics[width=\textwidth]{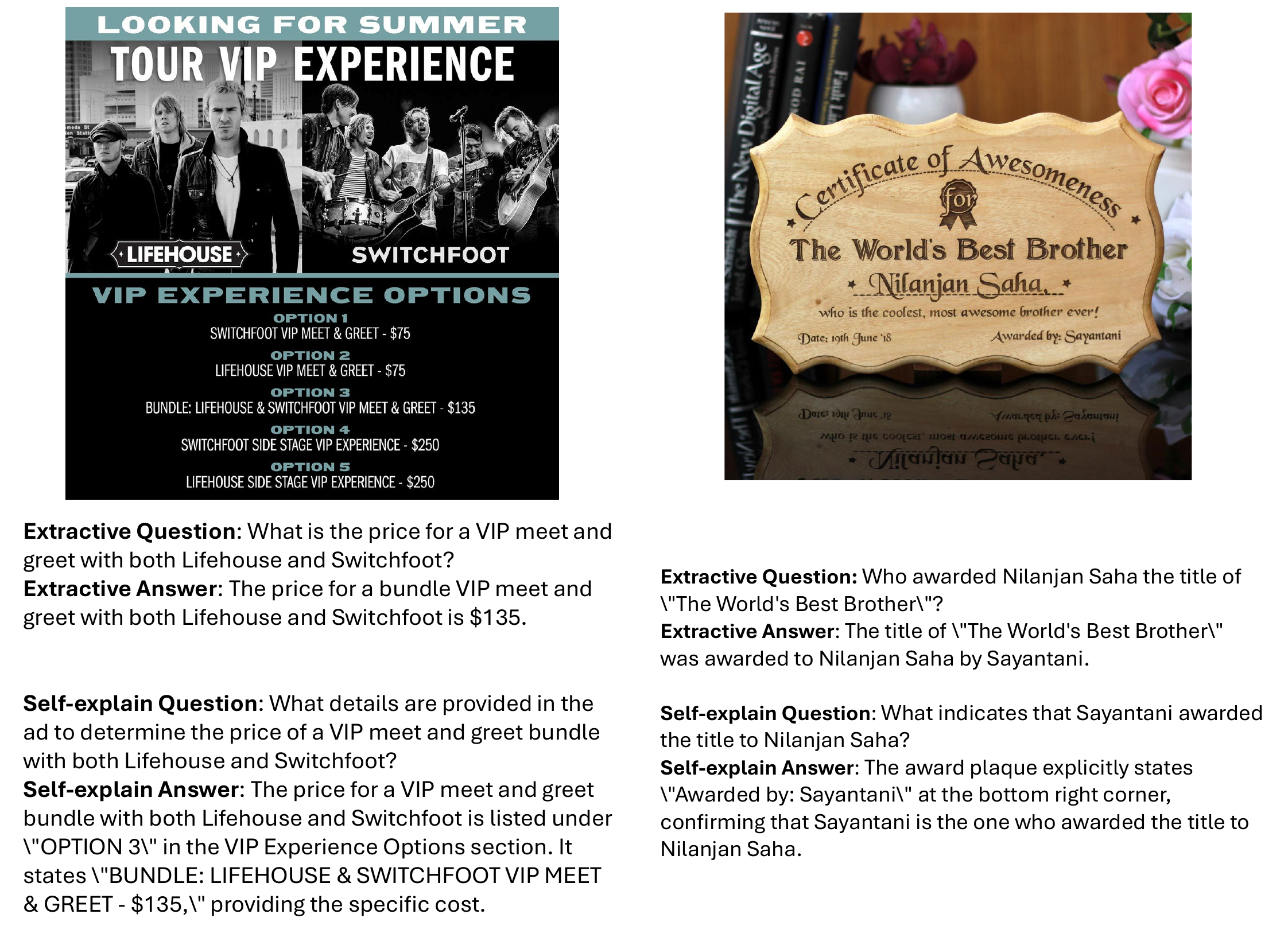}
    \hspace{3em}
    \includegraphics[width=\textwidth]{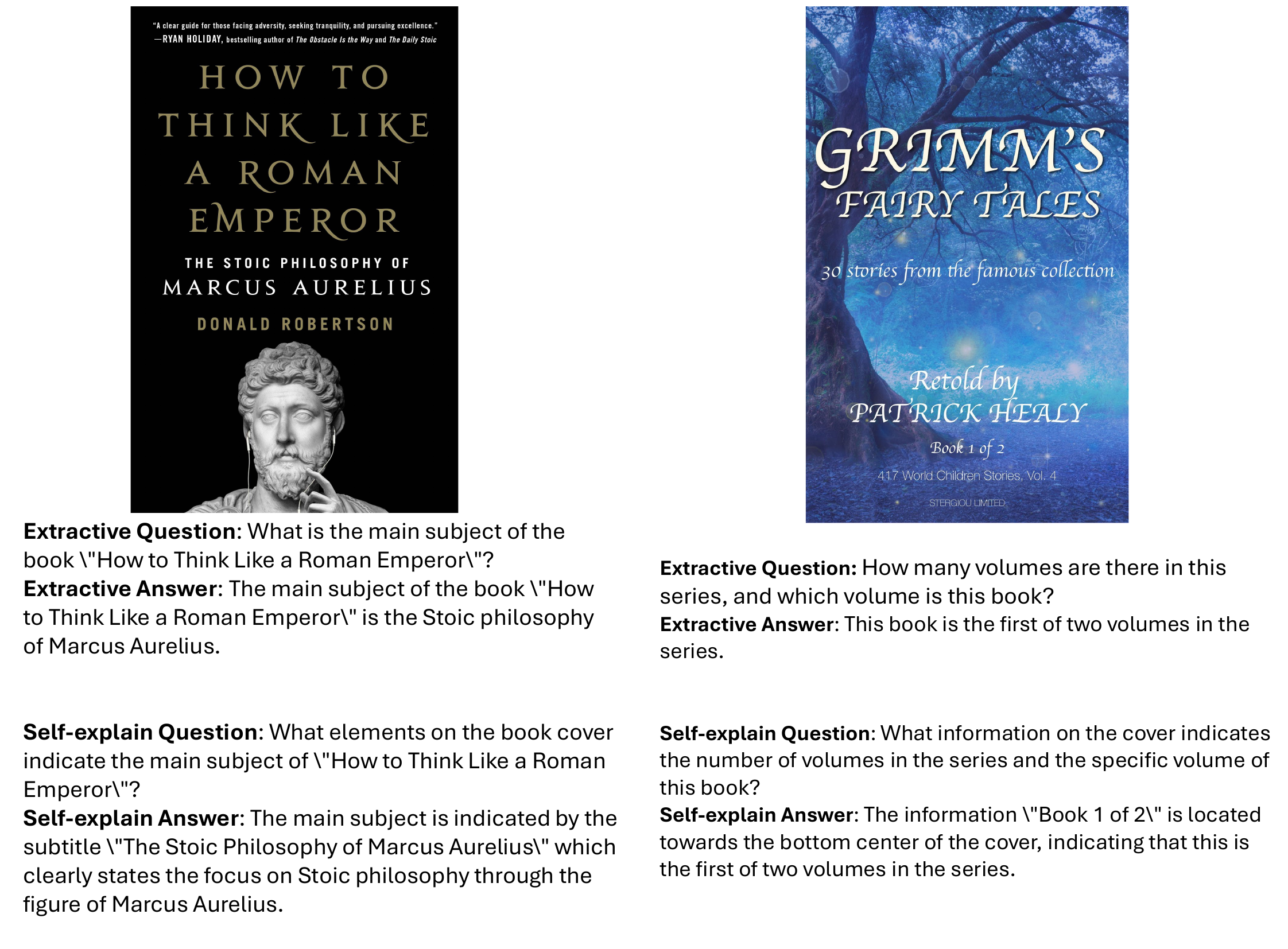}
    \caption{QA examples in LLaVAR-2-VQA}
    \label{fig:vqa1}
\end{figure*}

\begin{figure*}[!ht]
    \centering
    \hspace{-4mm}
    \includegraphics[width=\textwidth]{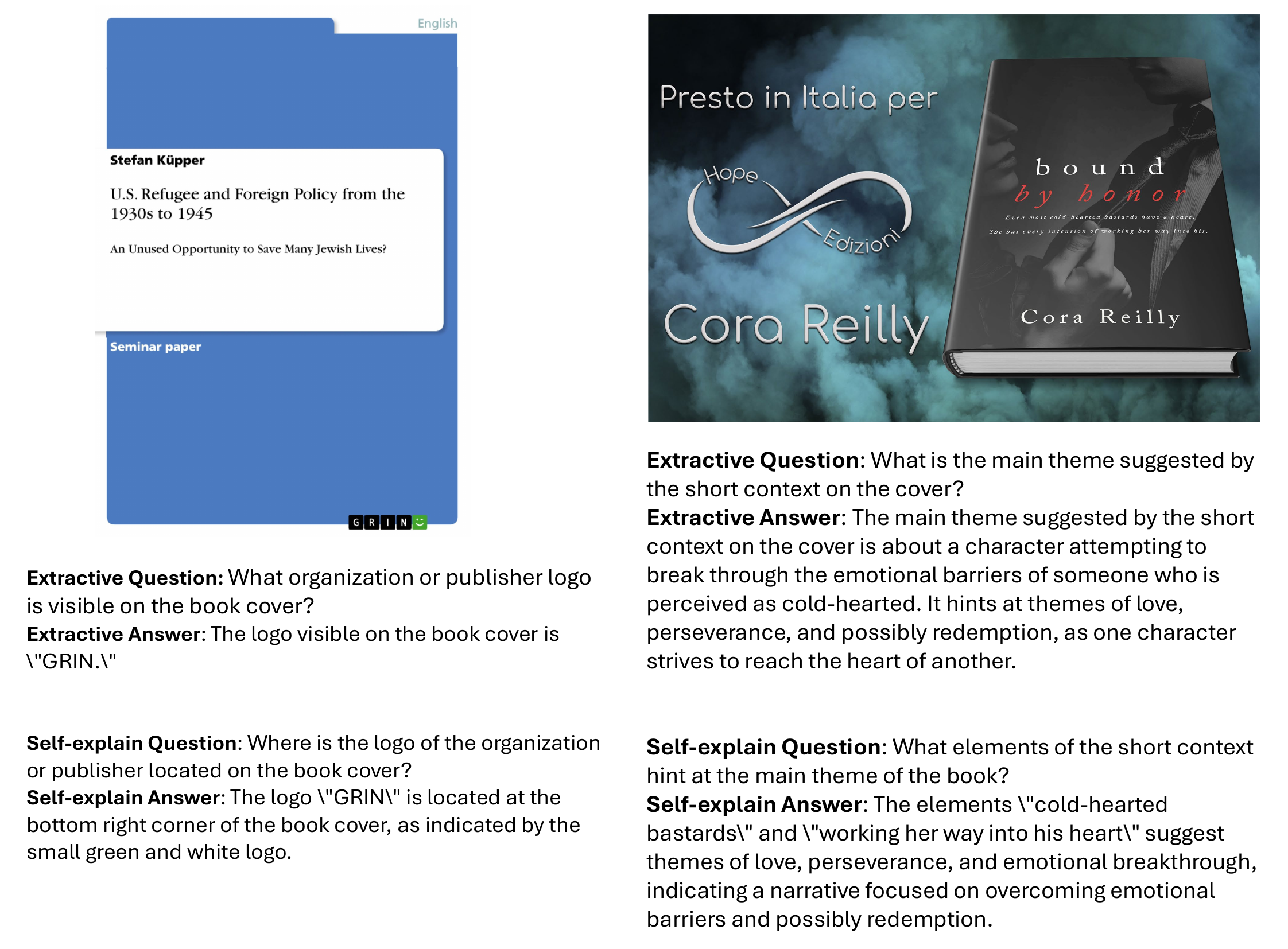}
    \caption{QA examples in LLaVAR-2-VQA}
    \label{fig:vqa2}
\end{figure*}

\begin{figure*}[!ht]
    \centering
    \hspace{-4mm}
    \includegraphics[width=\textwidth]{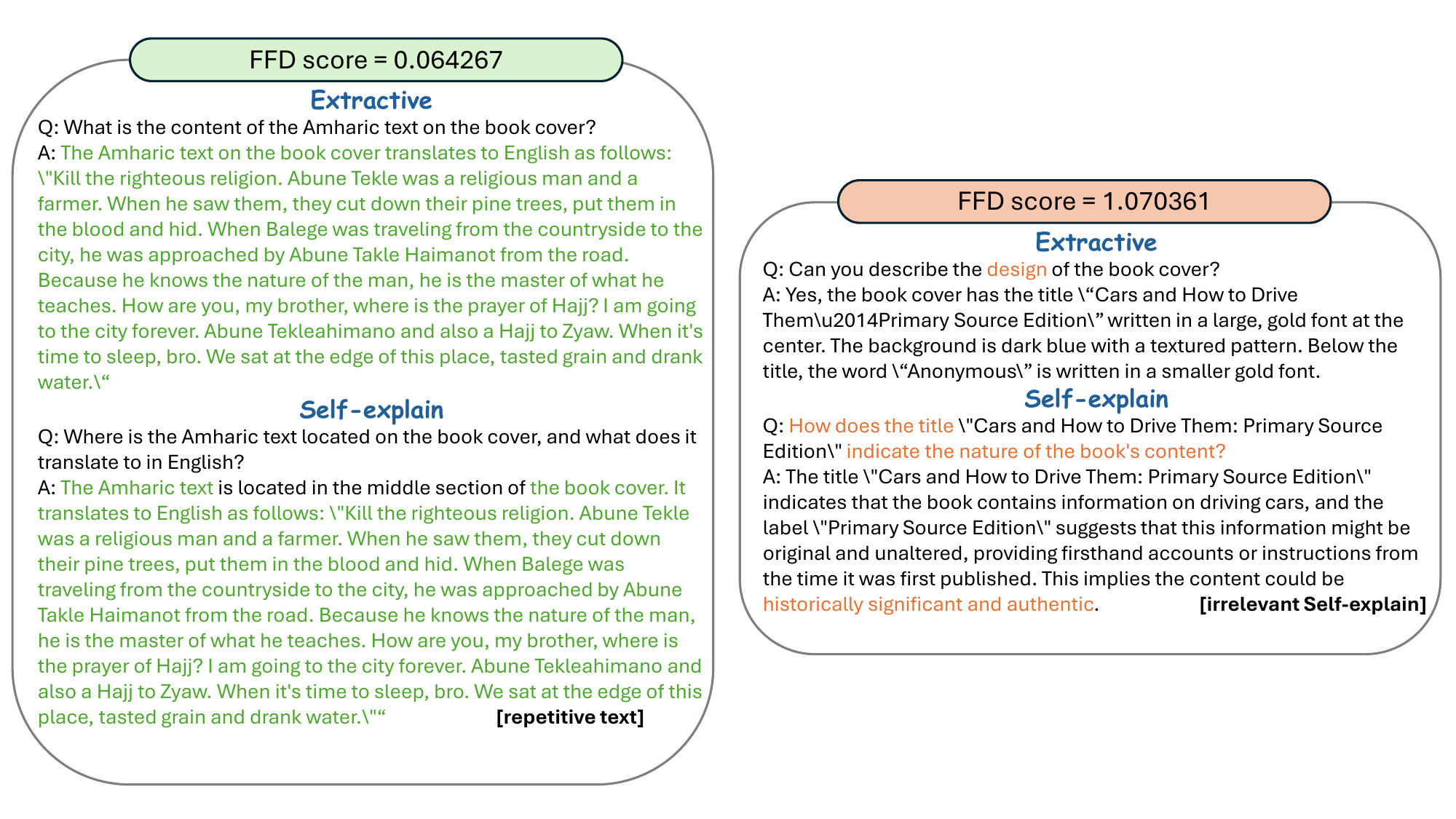}
    \caption{Filtered out Self-explain Examples based on FFD score}
    \label{fig:efd0}
\end{figure*}
\section{LLaVAR-2-VQA Examples} \label{appendix:vqa_example}
Figure~\ref{fig:vqa1} and Figure~\ref{fig:vqa2} present examples in LLaVAR-2-VQA.
\begin{figure*}[h!]
    \centering
    \hspace{-4mm}
    \includegraphics[width=\textwidth]{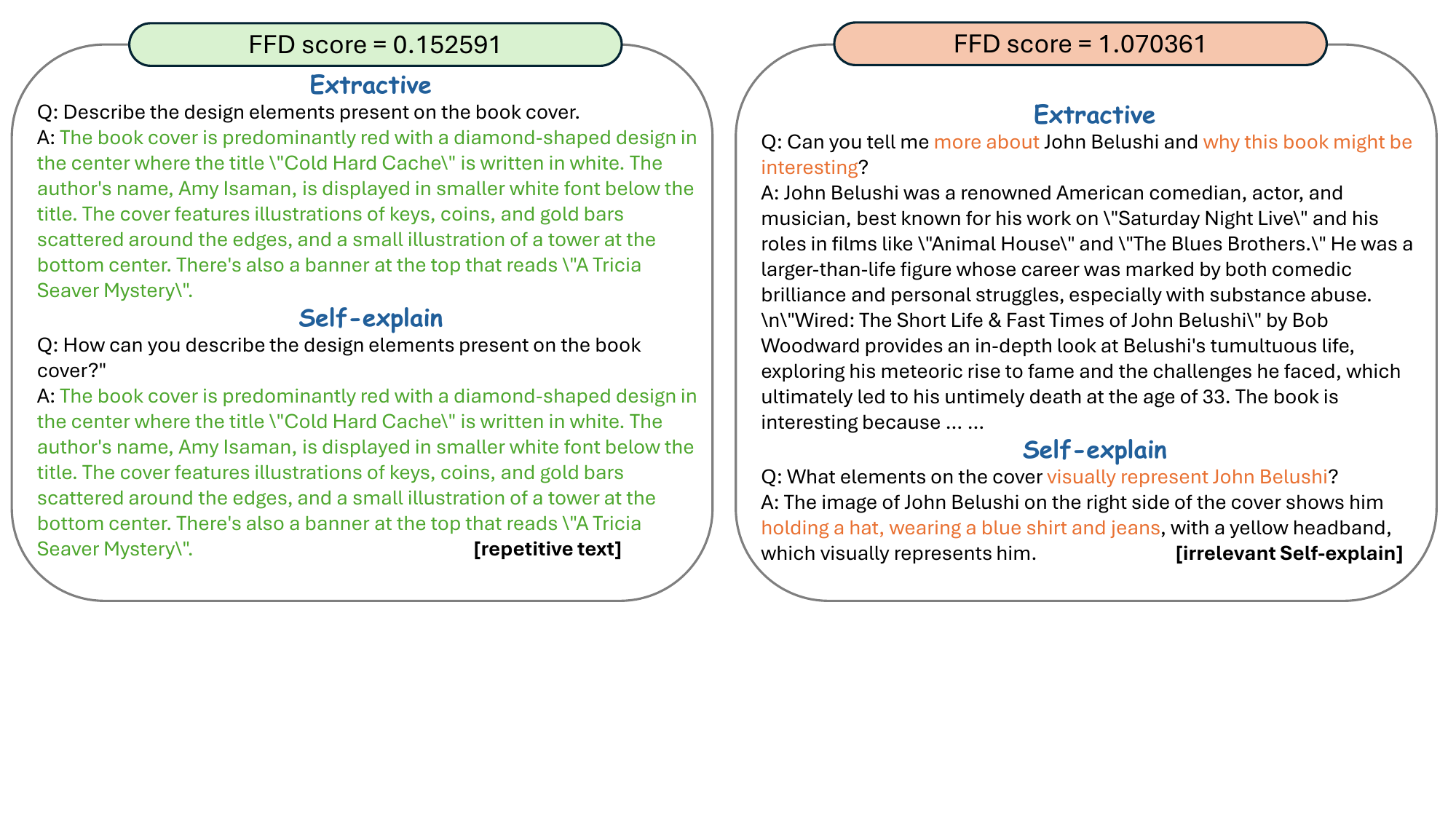}
    \caption{Self-explain Examples filtered out based on FFD score}
    \label{fig:efd1}
\end{figure*}
\begin{figure*}[h!]
    \centering
    \hspace{-4mm}
    \includegraphics[width=\textwidth]{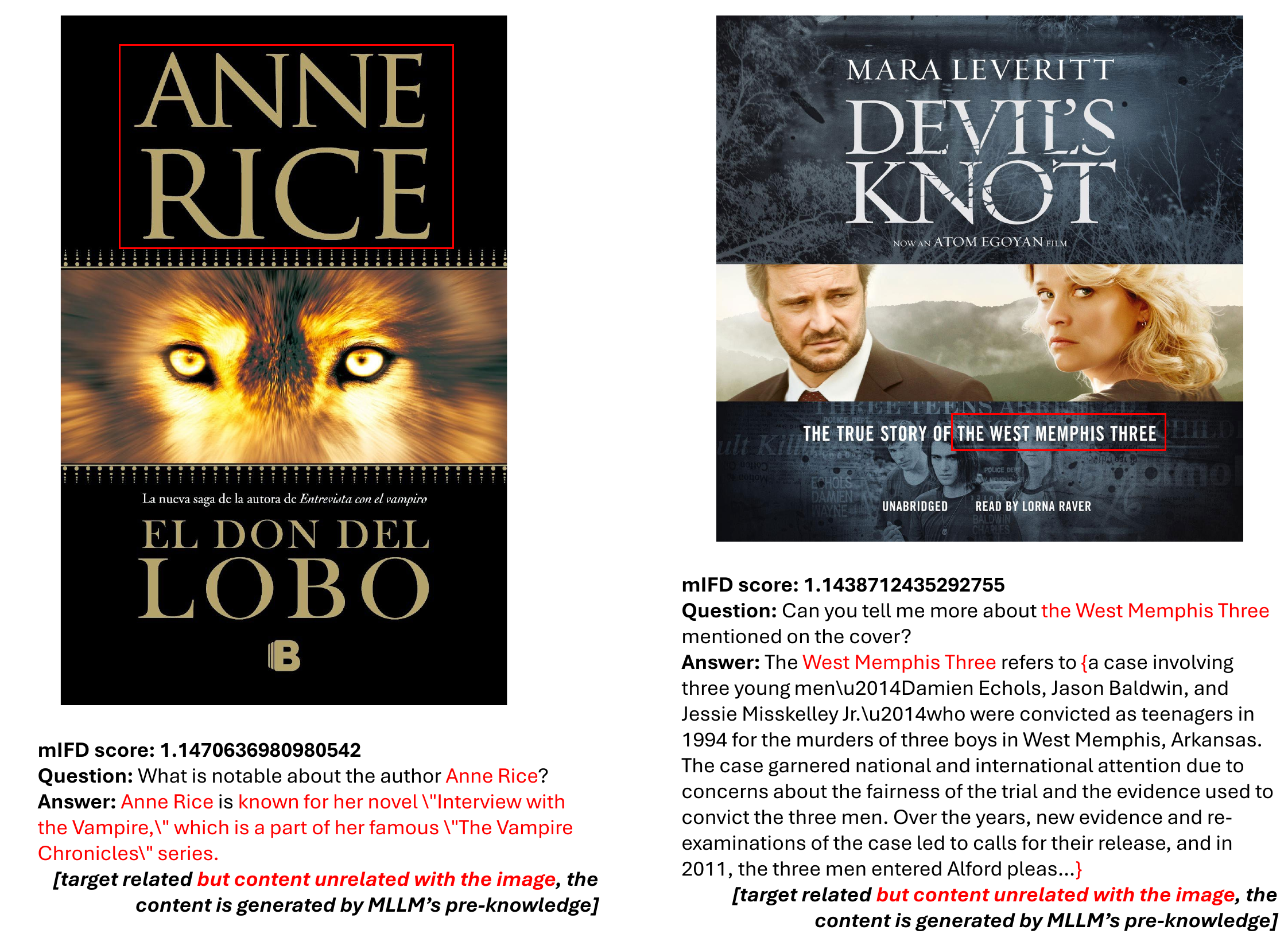}
    \hspace{3em}
    \includegraphics[width=\textwidth]{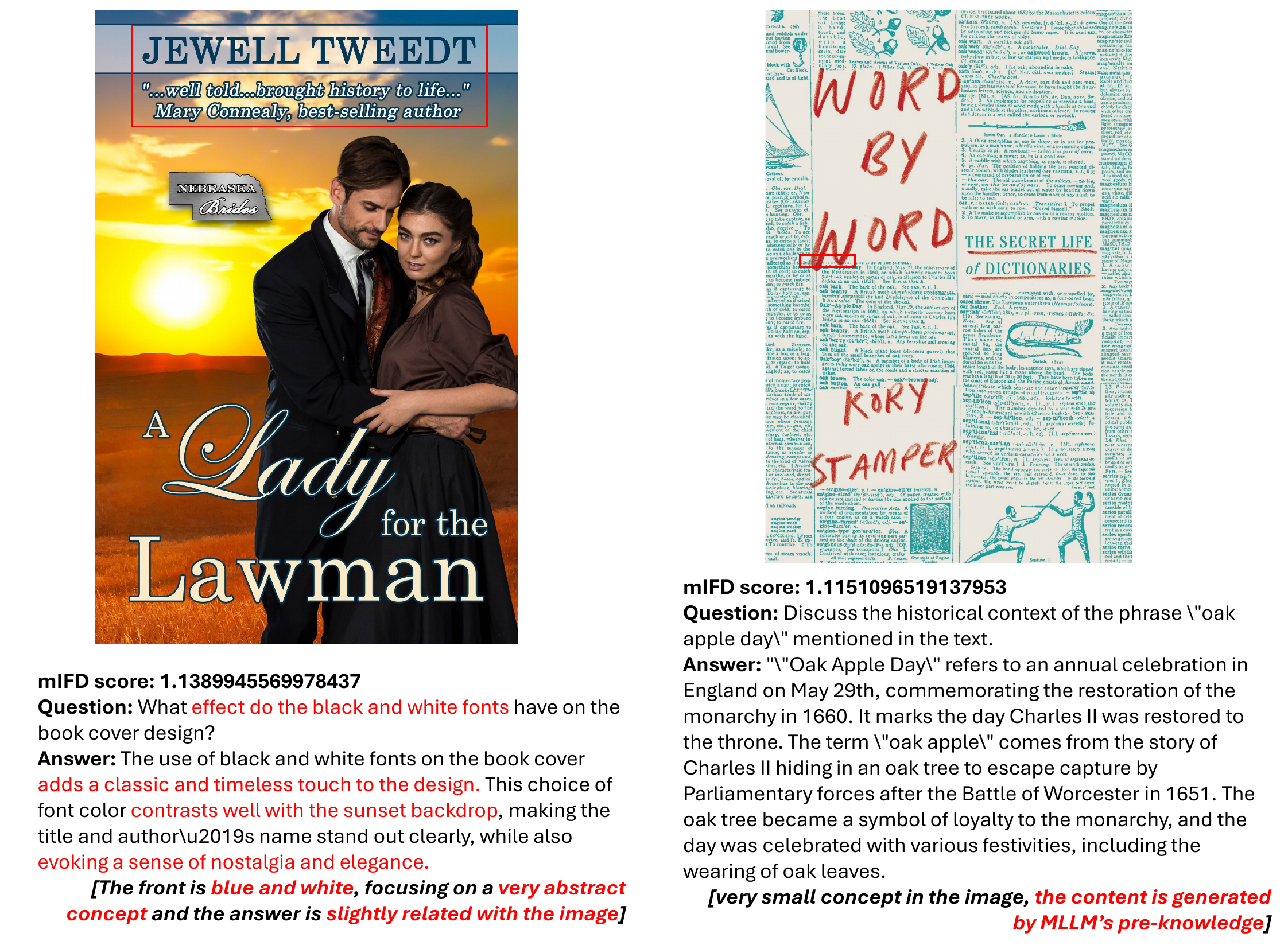}
    \caption{Extractive Examples filtered out based on mIFD score}
    \label{fig:mifd_example}
\end{figure*}
\section{Filtered out Examples}\label{sec:appendix_filter}
\paragraph{Extractive data filtered out by mIFD score} Figure~\ref{fig:mifd_example} presents extractive examples filtered out using mIFD score. The analysis is included in each example.
\paragraph{VQA examples filtered out by FFD score} Figure~\ref{fig:efd0} and Figure~\ref{fig:efd1} present examples filtered out using FFD score including 2 cases: FFD score is closed to 0 and FFD score is closed to 1.

\end{document}